\documentclass[letterpaper]{article} % DO NOT CHANGE THIS
\usepackage[preprint]{aaai2027}       % preprint: real authors, no copyright slug
\usepackage[hyphens]{url}   % DO NOT CHANGE THIS
\usepackage{graphicx}       % DO NOT CHANGE THIS
\usepackage{natbib}         % DO NOT CHANGE THIS AND DO NOT ADD OPTIONS
\usepackage{caption}        % DO NOT CHANGE THIS AND DO NOT ADD OPTIONS
\usepackage{amsmath}
\usepackage{amssymb}
\usepackage{booktabs}   % \toprule \midrule \bottomrule
\usepackage{multirow}   % appendix ablation table
\usepackage{array}

\newif\ifarxiv\arxivtrue

\title{SkillCorpus: Consolidating and Evaluating the Open Skill Ecosystem for Real-World LLM Agents}

\author{
    Yanze~Wang\textsuperscript{\rm 1,2,3}\equalcontrib,
    Pengfei~Yao\textsuperscript{\rm 1,2}\equalcontrib,
    Tianyi~Sun\textsuperscript{\rm 1,2,3},
    Chuanrui~Hu\textsuperscript{\rm 1,2}\footnote{Project leader.},
    Yan~Xiao\textsuperscript{\rm 1,2},
    Xiaotian~Luo\textsuperscript{\rm 1,2},
    Yunyun~Han\textsuperscript{\rm 1,2},
    Yifan~Chen\textsuperscript{\rm 1,2},
    Jun~Sun\textsuperscript{\rm 3}\corresponding,
    Yafeng~Deng\textsuperscript{\rm 1,2}\corresponding
}
\affiliations{
    \textsuperscript{\rm 1}EverMind\qquad
    \textsuperscript{\rm 2}Shanda Group\qquad
    \textsuperscript{\rm 3}Peking University\\
    \{yanze.wang, yaopengfei, tianyi.sun, chuanrui.hu, hanyunyun, yifan.chen, dengyafeng\}@shanda.com,\\
    \{yan.xiao, xiaotian.luo\}@evermind.ai, jsun@pku.edu.cn
}

\begin{document}
\maketitle

% =============================================================================
% Abstract
% =============================================================================
\begin{abstract}
Agent skills, \texttt{SKILL.md} files that package reusable
procedural knowledge for an LLM agent, are a popular
mechanism for extending agent capabilities.  Public
repositories now host them in large and growing numbers, yet these
artifacts are fragmented, redundant, and uneven in quality, and their
value in practice is unclear.  A core
question remains open, namely how to consolidate this open-source
\texttt{SKILL.md} ecosystem into a single usable corpus, and
what bounds its benefit on real-world agent tasks.
We present \textbf{SkillCorpus}, a framework that aggregates,
curates, matches, and evaluates the open skill ecosystem at
scale.  It filters $\sim$$821{,}000$ crawled skills through
a multi-stage pipeline into $96{,}401$ skills
organised by a 16-class taxonomy and three quality facets
(utility, robustness, safety), and pairs them with a fine-tuned
retrieval-and-selection stack that matches task-relevant skills.
We evaluate end-to-end across three benchmarks (SkillsBench,
GDPVal, QwenClawBench), two harnesses, and two open backbones
with a frontier robustness check.
Integrating SkillCorpus yields consistent gains across all
three benchmarks, largest on SkillsBench ($+7.5$\,pp).  An operational
analysis traces the gains to a coverage boundary and a harness
boundary.
SkillCorpus is, to our
knowledge, the first end-to-end account of when a curated,
retrieval-served community corpus improves real agent tasks, and
where it does not.
The dataset, models, and code are available at
\url{https://github.com/EverMind-AI/SkillCorpus}.
\end{abstract}

% =============================================================================
% Body — sections shared verbatim with the ACL working copy.
% =============================================================================
% =============================================================================
% §1 Introduction (~1.0 page) — rewritten as progressive narrative
% =============================================================================
\section{Introduction}
\label{sec:intro}

Large language models increasingly act as autonomous agents that
orchestrate long-horizon executable tasks rather than answer
isolated queries \citep{Yao2023ReAct,Shinn2023Reflexion,Park2023Generative}.
Yet on composite real-world workloads, even the strongest
frontier systems remain measurably bounded
\citep{Merrill2026TerminalBench,OpenAIGDPVal2025}.  This
gap is difficult to close by raw model scaling alone.  A growing
consensus instead treats agent capability as a layered system, in
which a frozen foundation model is paired with externalised
skills and an agent harness that loads and executes them
\citep{AnthropicSkillCreator2025}.  A skill
packages reusable procedural knowledge invoked at inference time.
Unlike opaque, slow-to-update model weights or transient
in-context examples, it is interpretable and modular.  
The same externalise-and-retrieve pattern is already well-established for tool use,
where agents retrieve and invoke calls over large API inventories
\citep{Qin2024ToolLLM,Patil2023Gorilla} and the recent Model Context Protocol (MCP) further standardises access to external tools and resources \citep{Hou2025MCP,MCPCorpus2025}.
Skills extend this pattern from external tool calls to
procedural knowledge itself.

Industry-grade harnesses now ingest skill files routinely
\citep{AnthropicClaudeCode2025,OpenAIClaudeCode2025}, and a
community ecosystem has formed around the \texttt{SKILL.md}
specification \citep{AnthropicSkillCreator2025}, now spanning
thousands of public repositories.  Whether this abundance actually improves agent performance on real tasks, and what it takes to make it do so, remains
unsettled.  Curated skills paired one-to-one with tasks raise pass
rates on average but with sharply heterogeneous per-domain effects
\citep{SkillsBench2026}, while uncurated community libraries can
fail to improve over a no-skill baseline
\citep{SkillFlow2026,UCSBSkillsWild2026}.  Two gaps stand in the
way.  \textbf{First}, we know of no released corpus that is at once broadly
aggregated, strictly quality-gated, and cleared for open
redistribution.  Existing
artifacts draw from only one or two source channels~\citep{SkillFlow2026,UCSBSkillsWild2026}, and those that filter for quality show no evidence of broad coverage at
scale~\citep{SkillNet2026}.  
\textbf{Second}, the downstream value of a community corpus on real agent tasks is largely uncharacterised. Prior evaluations pair tasks with hand-picked oracle skills rather than a deployed corpus~\citep{SkillsBench2026}, run in simulated environments~\citep{SkillNet2026}, or measure in-the-wild usage through a single harness over limited sources~\citep{UCSBSkillsWild2026}.

% Closing both gaps calls for four capabilities at once.  Broad
% aggregation and strict curation address the resource side.
% Precise matching and end-to-end evaluation address the
% measurement side, since a fair test must retrieve skills from the
% corpus rather than hand-pick an oracle.  We present
% \textbf{SkillCorpus}, which delivers all four and closes both gaps.
To close both gaps, we present \textbf{SkillCorpus}, a single end-to-end framework that consolidates the open \texttt{SKILL.md} ecosystem into a curated, deployable corpus and evaluates it on real-world agent tasks — making it possible to measure when community skills help real agents, and where they do not.

Our contributions are threefold:
\begin{itemize}
\item \textbf{Framework.} We unify aggregation, curation, matching, and real-world evaluation of the open \texttt{SKILL.md} ecosystem, turning a fragmented, redundant, and uneven pool of community skills into a curated foundation for reuse at scale.

\item \textbf{Open resource.} We release the resulting $96{,}401$-skill corpus (curated from $\sim$$821{,}000$ raw files), its fine-tuned retrieval-and-selection stack, and the full curation, training, and evaluation code, all licence-audited and $100\%$ OSI-permissive — a fully open, ecosystem-scale resource for skill-augmented agent research.\ifarxiv\footnote{Code, data, and models: \url{https://github.com/EverMind-AI/SkillCorpus}.}\fi

\item \textbf{Findings.} Evaluating across three real-world agent benchmarks, independent harnesses, and open-to-frontier backbones (\S\ref{sec:experiments}), we find that integrating SkillCorpus consistently improves agent performance, with gains modulated by corpus coverage and the harness, characterising the real utility and limits of community skills.

%\item We propose SkillCorpus, a single end-to-end framework that
%      unifies all four capabilities over the open \texttt{SKILL.md}
%      ecosystem, turning a scattered, uneven pool of community skills
%      into a curated foundation for reuse at scale.

% \item We release the resulting $96{,}401$-skill corpus curated from
%       $\sim$$821{,}000$ raw files, its fine-tuned
%       retrieval-and-selection stack, and the full curation, training,
%       and evaluation code, all $100\%$ OSI-permissive,
%       an end-to-end open, ecosystem-scale resource for skill-augmented
%       agent research.

% \item Evaluating across multiple real-world agent benchmarks,
%       independent harnesses, and open-to-frontier backbones
%       (\S\ref{sec:experiments}), we find that integrating SkillCorpus
%       consistently improves agent performance, with gains modulated by
%       corpus coverage and the harness, characterising the real utility
%       and limits of community skills.
\end{itemize}

% =============================================================================
% §2 Related Work (~0.7 page) — narrative, minimal redundancy with §1
% =============================================================================
\section{Related Work}
\label{sec:related}

\subsection{Skill acquisition and curation}

Work on the skill layer divides by when skills enter
the system.  One strand acquires them at run time, while the
agent is solving tasks.  Voyager grows a skill library through
environment exploration \citep{Wang2023Voyager}, Toolformer
trains the model to invoke tools self-supervisedly
\citep{Schick2023Toolformer}, ToolLLM and Gorilla scale API
calling to large tool inventories
\citep{Qin2024ToolLLM,Patil2023Gorilla}, and recent
agent-evolution work lets agents redesign their own scaffolding
\citep{Zhang2024Memento,AEvolve2025}.  All of these build on
agent harness scaffolds \citep{Yao2023ReAct,Shinn2023Reflexion}.
The other strand, which this paper occupies, prepares skills
before the agent ever runs.  We curate a
community-contributed corpus at ingest time, so that runtime
systems have a trustworthy library to draw from.  This
ingest-time curation mirrors data-quality work in language-model
training, where deduplication \citep{Lee2022Dedup} and
quality-based data selection \citep{Zhou2023LIMA} improve
downstream models more than raw scale.

Released community skill corpora vary in emphasis.
\citet{SkillNet2026}, concurrent with our work, releases a
200K-skill ontology rated along five quality rubrics and linked
by a skill-relation graph.  It measures benefit on
simulated environments, with skills synthesised from each
benchmark's own expert trajectories rather than on real tasks
served by a deployed corpus.  \citet{SkillFlow2026} pairs a 36K
uncurated library with a four-stage retrieval pipeline and
treats quality only implicitly, through artifact-density
signals.
\citet{AgentSkillsDataAnalysis2026} provides a data-driven
characterisation of a smaller curated set.  None of these releases
reports a source-licence audit, so their redistribution rights are
left unresolved.  Nor does prior work tie these per-skill scores to
their downstream operational role.  We introduce a
compact three-facet framework chosen for measurable facet
separation rather than rubric coverage, and characterise its
operational signal directly (\S\ref{sec:claim-a}).

\subsection{Skill retrieval and evaluation}

Skill retrieval architectures range from lexical and dense
recall to learned skill routers \citep{SkillRouter2026},
supported by dedicated retrieval benchmarks \citep{SkillRet2026},
corpus-size scaling analyses \citep{ScalingLaws2026}, and
dependency-aware composition \citep{GraphOfSkills2026}.  A
complementary line of work emphasises skill orchestration and
context-injection over retrieval per se.
\citet{AgentSkillOS2026} couples capability-tree retrieval with
DAG-based orchestration at $\sim$$200$K scale, and
\citet{SkillRAE2026} reports $+11.7\%$ on SkillsBench via a
skill graph plus compact compilation.  That gain, however, is
measured against the prior state-of-the-art injection method
rather than a no-skill or community-corpus baseline, so it is not
directly comparable to our end-to-end $\Delta$.
The retrieval strand reports ranking metrics on its
own curated pools rather than task outcomes.  Our
stack is fine-tuned on the deduplicated
corpus it serves and adds an LLM selector that reads full skill
bodies before injection, with both its standalone retrieval
performance and its end-to-end effect on task pass rates
measured directly.

On evaluation, \citet{SkillsBench2026} characterises oracle-skill
efficacy, and \citet{UCSBSkillsWild2026} reports diminishing
returns on uncurated libraries at scale.  Neither
evaluates a single community-curated corpus under a
deployable retrieval-and-selection stack across multiple
real-world benchmarks.  \S\ref{sec:experiments} fills this gap
with a cross-benchmark, cross-harness, multi-backbone measurement
of end-to-end task gains.

% =============================================================================
% §3 SkillCorpus (~1.5 pages)
% =============================================================================
\section{SkillCorpus}
\label{sec:corpus}

% FIG 1 (fig:pipeline) declared at the top of §3 (method) so the full-width [t]
% float lands on the method page. AAAI bans stfloats/dblfloatfix, so placement
% is controlled by the declaration point.
\begin{figure*}[t]
\centering
\includegraphics[width=\textwidth]{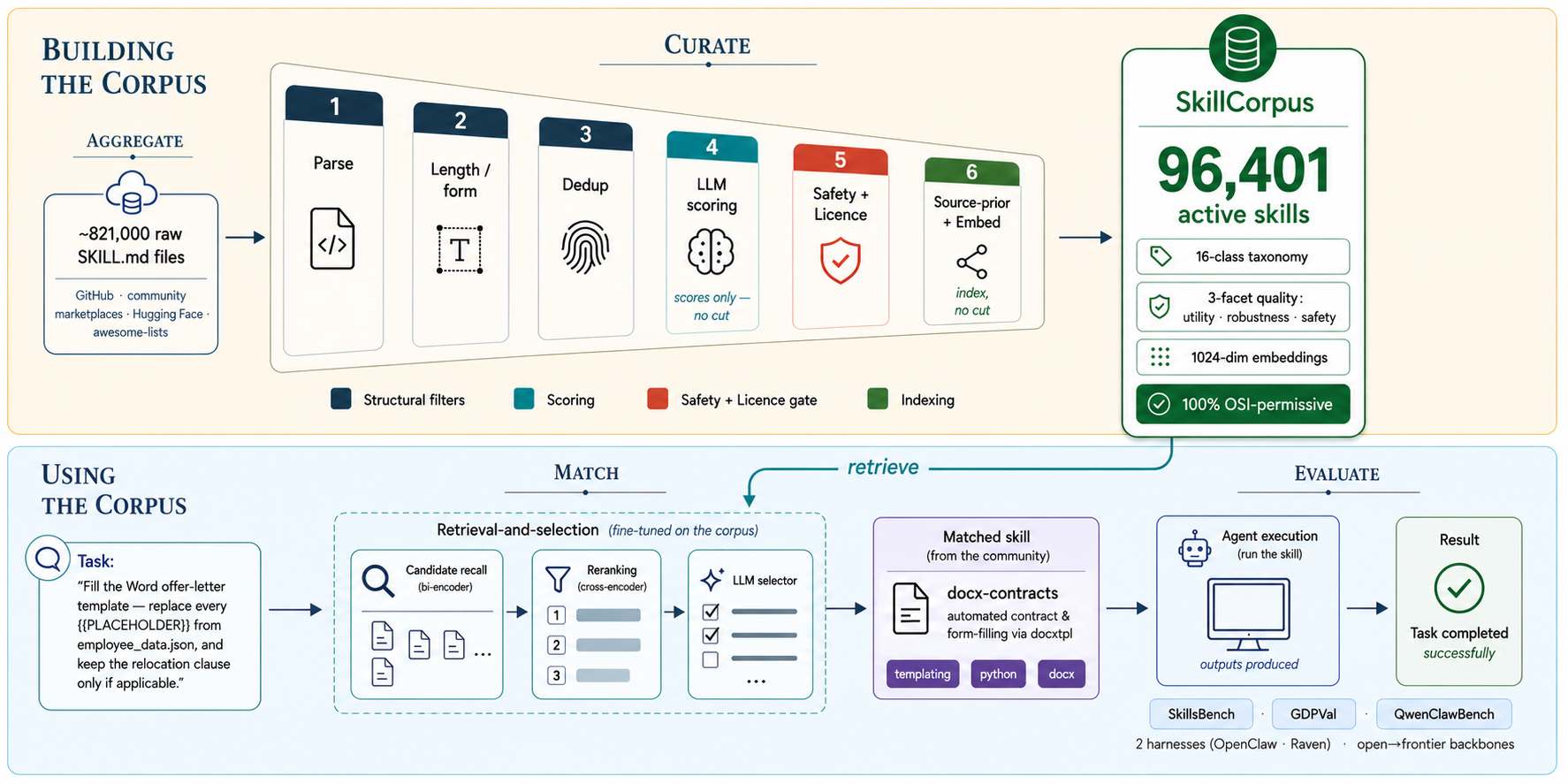}
\caption{The SkillCorpus framework.  Building the corpus (top), we
aggregate community \texttt{SKILL.md} files from public source
channels and curate them through a six-stage funnel into a
$96{,}401$-skill corpus organised by a 16-class taxonomy and three
quality facets.  Using the corpus (bottom), a fine-tuned
retrieval-and-selection stack matches skills to an incoming task and
an agent runs the selected skill end-to-end.  The worked example
traces one real task from request to result, and we evaluate this
pipeline across three real-world benchmarks, two harnesses, and
open-to-frontier backbones.}
\label{fig:pipeline}
\end{figure*}

% \paragraph{Skill specification.}
% We adopt the \texttt{SKILL.md} convention of
% \citet{AnthropicSkillCreator2025} verbatim.  Each skill is a
% self-contained directory whose \texttt{SKILL.md} file carries
% required YAML frontmatter (\texttt{name}, \texttt{description})
% plus a Markdown procedural body, optionally bundled with
% executable resources.  The ingest pipeline below admits only
% files conforming to this contract, matching the loading
% convention used by the agent harnesses in
% \S\ref{sec:experiments}.

\ifarxiv
SkillCorpus is a single end-to-end framework that turns the open \texttt{SKILL.md} ecosystem into a curated, deployable corpus and measures its effect on real-world agent tasks. As shown in Figure~1, it runs in four stages: aggregation and curation of the ecosystem into a corpus (\S3.1–\S3.3), matching of task-relevant skills via a fine-tuned retrieval-and-selection stack (\S3.4), and end-to-end evaluation on real-world agent tasks (\S\ref{sec:experiments}).
\else
As shown in Figure~1, SkillCorpus runs in four stages: aggregation and curation of the ecosystem into a corpus (\S3.1–\S3.3), matching of task-relevant skills via a fine-tuned retrieval-and-selection stack (\S3.4), and end-to-end evaluation on real-world agent tasks (\S\ref{sec:experiments}).
\fi

% --------------------------------------------------------- 3.1 pipeline ------
\subsection{Multi-source aggregation and the curation pipeline}
\label{sec:pipeline}

The corpus's value is its coverage, spanning the whole public
ecosystem rather than a single channel.
The crawl is driven by a single machine-readable source
registry of $62$ sources, detailed in
Appendix~\ref{app:crawl-sources}.  The registry spans five
ingestion mechanisms, from directly cloned repositories and
awesome-list scrapes to marketplace index APIs, sitemap crawls,
and JSON catalogs, and every mechanism feeds one shared discovery
entry point.  Our crawl gathers $\sim$$821{,}000$ raw \texttt{SKILL.md} files.
After source selection and repository-level de-duplication,
$25{,}159$ GitHub repositories feed the six-stage pipeline, which
curates them into the released active set.

Figure~\ref{fig:pipeline} visualises the pipeline.
% Stages~1--2 apply structural filters on the
% \texttt{SKILL.md} contract (parse and length/form).  
Each skill follows the \texttt{SKILL.md} convention of \citet{AnthropicSkillCreator2025} — a Markdown body with optional executable resources; Stages 1–2 apply structural filters on this contract (parse and length/form).
Stage~3 deduplicates.  Stage~4 runs the LLM judge that emits the
3-facet quality score and 19 safety/quality flags.  Stage~5
applies the safety hard-gate plus
Open Source Initiative (OSI)-permissive licence filter
(Table~\ref{tab:license-filter}).  Stage~6 attaches source-prior
shrinkage, the 1024-dim retrieval embedding, and index entries.
Parse and deduplication dominate the funnel, while the remaining
filters contribute a few percent.

\paragraph{Two-tier deduplication (stage 3).}
Deduplication is the funnel's steepest reduction
($283{,}844 \to 101{,}111$, $-64\%$), in two tiers.  An exact tier
collapses $169{,}465$ content- and name-fingerprint duplicates,
$59.7\%$ of the stage input, which reflects how
heavily the ecosystem replicates the same artifacts across
repositories.  A semantic tier embeds each survivor (1024-dim)
and flags pairs above cosine $0.90$, and pairs above $0.995$
merge automatically.  The $66{,}751$ borderline pairs are adjudicated
by an LLM judge, which confirms $25.4\%$ of pairs and collapses
$13{,}268$ near-duplicate skills.  The resulting low-overlap
distribution also lets the retrieval stack train on the corpus
itself.

% --------------------------------------------------------- 3.2 quality -------
\subsection{Three-facet quality framework}
\label{sec:quality}

\ifarxiv
We decompose skill quality into three
facets, each capturing a distinct concern, rather than
collapsing them into a single composite or a flat
multi-dimensional list.
\else
We decompose skill quality into three facets, each capturing a
distinct concern rather than a single composite.
\fi
These facets serve curation and
retrieval ranking, with the safety facet additionally serving the
release gate, as characterised in \S\ref{sec:claim-a}.

\paragraph{Three facets.}
The utility facet scores the description alone, asking whether the
task class is well-scoped with clear triggers and applicability
conditions.  The robustness facet scores the body content and its
consistency with the description's promise.  A body that is
internally coherent but silently delivers something narrower than
promised is penalised here.  The safety facet captures harm and risk along
categories such as prompt injection, command injection,
credential leakage, and unsafe execution.  By design, the utility and robustness facets are scored
independently.  Description-level issues only affect utility, and
body-level issues only affect robustness.

\paragraph{Flag taxonomy.}
The 19-flag vocabulary is partitioned across the three facets ($2$
utility-bound, $6$ robustness-bound, $11$ safety-bound), with the
full list in Appendix~\ref{app:flags}.  Each flag, when fired, constrains
its assigned facet's score.  Five flags act as hard gates:
\texttt{prompt\_injection}, \texttt{cmd\_injection},
\texttt{unsafe\_exec}, \texttt{auth\_bypass}, and
\texttt{csam\_risk}.  Firing any one of these forces the quality
score to zero and excludes the skill from the released
\texttt{active} set.  The remaining fourteen flags act as soft signals.  When such a
flag fires, the LLM judge caps the corresponding facet score.
A fired \texttt{placeholder} flag, for example, holds robustness
at $\leq 4$.  No additional numeric deduction is applied, so the same
observation is not double-counted.

\paragraph{Facet separation.}
The three facets are not redundant.  The safety facet is markedly
more independent from the two craftsmanship facets
($r(u,s){=}0.15$, $r(r,s){=}0.30$) than those are from each other
($r(u,r){=}0.59$), matching the design, since $u$ and $r$ both
track artifact craftsmanship while $s$ targets an orthogonal harm
axis on which a well-crafted skill can still be unsafe.  We read
these contrasts qualitatively and do not claim construct-validity
orthogonality.  Rigorous validation would require independent human
gold annotation \citep{SkillNet2026}.  The off-diagonal-mass
analysis, the per-pair figure, and the score-bucket distribution
are given in Appendix~\ref{app:quality-detail}.

\paragraph{Score formula.}
Following the LLM-as-judge paradigm \citep{Zheng2023LLMJudge},
the judge outputs three integer subscores $u, r, s$ on
$\{0, \ldots, 10\}$, normalised to $[0, 1]$.  These combine into
$\mathtt{content\_q} = 0.50\,u + 0.35\,r + 0.15\,s$, attenuated by
a factor $0.5 + 0.5\,(s{-}0.3)/0.4$ when the safety subscore is
marginal ($s \in [0.3, 0.7]$).  The final per-skill score, or composite quality score, shrinks
content quality toward a per-source prior
(Appendix~\ref{app:source-prior}):
\[
\mathtt{score} = 0.85\,\mathtt{content\_q} + 0.15\,\mathtt{prior}_{\mathrm{src}} + b,
\]
where $b$ is a small structural bonus
($+0.05$ for skills shipping \texttt{scripts/},
$+0.02$ for \texttt{references/}), and the result is clipped to
$[0, 1]$.

% --------------------------------------------------------- 3.3 classify -----
\subsection{Classification, filtering, and release}
\label{sec:classify}

\paragraph{Taxonomy.}
We adopt a single-label, 16-class taxonomy designed to cover the
observed distribution of community skills while remaining
decision-tree classifiable from \texttt{name},
\texttt{description}, and a short body excerpt.  The classifier
prompt enforces six conflict-resolution rules, the most
consequential being the \textsc{AI-ML} rule that a skill is
labelled \textsc{AI-ML} only when its primary deliverable is an AI
system (persona, agent, or model), not merely when it uses an LLM
internally.  The full rule set is released with the classifier
prompt.

\begin{figure}[!ht]
\centering
\includegraphics[width=0.96\columnwidth]{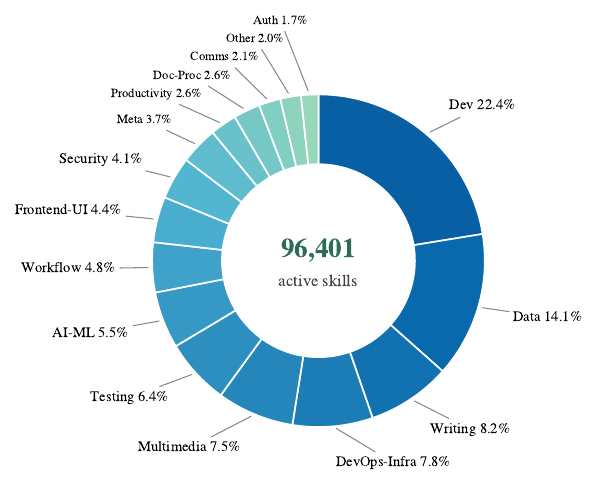}
\caption{16-class distribution on the active release set
($N{=}96{,}401$, shares in \%, sorted clockwise).}
\label{fig:class-dist}
\end{figure}

\paragraph{Distribution.}
The class distribution on the active release set is summarised in
Figure~\ref{fig:class-dist}.  The corpus is software- and
data-leaning.  \textsc{Dev}, \textsc{Data}, \textsc{Writing}, and
\textsc{DevOps-Infra} together account for over half of active
skills.  The catch-all \textsc{Other} class is small ($2.01\%$),
indicating that the taxonomy covers the observed distribution well.
Every active skill carries exactly one of the sixteen
classes.

\paragraph{Safety and licence filters (stage 5).}
The two filters run in sequence on the
$\sim$$101$K skills surviving stages 1--4.  A safety hard-gate
removes $915$ skills triggering any of three
conditions: a pre-judge \texttt{blocked.malware} regex match
(Appendix~\ref{app:fp-audit}),
firing of any of the five LLM-judged hard-gate flags defined in
\S\ref{sec:quality}, or an LLM safety subscore below~$3$.  A
licence filter then enriches each source repository's
\texttt{spdx\_id} via the GitHub API and admits only skills
carrying an OSI-approved permissive licence
(Table~\ref{tab:license}), excluding a further $3{,}795$ skills
whose source repositories lack a usable permissive licence
(per-reason counts in Table~\ref{tab:license-filter}).  Every
skill in the resulting $96{,}401$-skill active set is therefore
commercially redistributable under its source repository's
declared licence.

\paragraph{Release.}
The corpus is released as a single artifact comprising the
metadata SQLite database, the per-stage
ingest report of raw counts at each pipeline stage, the precomputed
3-facet quality cache, the 1024-dim retrieval embeddings, and
the retrieval-and-selection stack
used in our experiments and released as engineering tooling.  All
curation source code will be released under a permissive licence
upon acceptance.

% --------------------------------------------------------- 3.4 retrieval ----
\subsection{Retrieval and selection stack}
\label{sec:retrieval}

The corpus is released together with a fine-tuned
retrieval-and-selection stack so the artifact ships ready to use
rather than requiring a retrieval stack to be built.  We use ``deployable'' in this packaging sense, not as a
production-validation claim.  The stack follows the standard
retrieve-then-rerank pattern of retrieval-augmented generation
\citep{Lewis2020RAG}, with three operational stages: candidate
recall, rerank, and final selection by an LLM gate.

\paragraph{Recall and rerank.}
Candidate recall uses an embedding model fine-tuned from
Qwen3-Emb-0.6B.  A scoring model fine-tuned from Qwen3-Rank-0.6B
then reranks the top candidates.  Both models are trained on
the strictly deduplicated active set.  The low-overlap data
distribution is intended to give a cleaner contrastive learning
signal by preventing functionally identical near-duplicates from
being sampled as hard negatives, which could otherwise produce
false-negative collisions during contrastive fine-tuning.  We
adopt this as a training-design rationale rather than an isolated
ablation.  Recall and rerank operate over a
$3{,}000$-character retrieval field distilled from each skill
body, avoiding length-induced truncation noise.

\paragraph{LLM selector (gate).}
Reranked top-$k$ candidates can still include topically
irrelevant items, and brief summaries are insufficient for an
agent to judge tool fit at execution time.  We therefore insert
an LLM selector that, given the task query and the full skill
body of each reranked candidate, returns $0$--$2$ precise tools
for injection into the agent prompt.  This trades one additional
LLM call for the reliability of full-body inspection while
keeping the injected context within strict per-task budgets.

\paragraph{Query rewriter (optional).}
An optional LLM-based query rewriter normalises domain-specific
jargon in the task description to vocabulary that better matches
the corpus's skill descriptions, and is permitted to short-circuit
(no skills required) when the task is clearly outside the
corpus's scope.

We evaluate this stack both as part of the end-to-end pipeline,
where it drives the main-grid pass-rate gains, and as a
standalone retrieval system on Hit@1 / Recall@10 against
published baselines.

% =============================================================================
% §4 Experiments and Results (~3.0 pages)
% =============================================================================
\section{Experiments and Results}
\label{sec:experiments}

% =============================================================================
\subsection{Setup}
\label{sec:setup}

% Table 1 (tab:main-eff) declared inside §4.1 so it is queued while the Setup
% page composes and its full-width table*[t] float lands at the top of the next
% page, alongside the §4.2 Main results discussion (dblfloatfix).
\begin{table*}[!t]
\centering\small
\setlength{\tabcolsep}{7pt}
\begin{tabular}{@{}lccc@{}}
\toprule
 & SkillsBench & GDPVal & QwenClawBench \\
Cell & \scriptsize no-skill$\to$skill & \scriptsize no-skill$\to$skill & \scriptsize no-skill$\to$skill \\
\midrule
OpenClaw $\times$ Q-27B    & $8.8 \to 13.0$  & $81.2 \to 83.1$ & $65.2 \to 66.7$ \\
OpenClaw $\times$ Q-397B   & $11.1 \to 16.9$ & $82.2 \to 84.0$ & $65.7 \to 67.0$ \\
Raven $\times$ Q-27B       & $10.0 \to 16.5$ & $82.6 \to 83.8$ & $66.9 \to 70.8$ \\
Raven $\times$ Q-397B      & $9.2 \to \mathbf{22.6}$ & $84.0 \to 85.2$ & $68.8 \to 73.2$ \\
\midrule
Pooled $\Delta$ & $+7.5{\pm}2.3$ ($z{=}3.2$) & $+1.51{\pm}0.49$ ($z{=}3.1$) & $+2.79{\pm}0.70$ ($z{=}4.0$) \\
\bottomrule
\end{tabular}
\caption{Main results: absolute performance per cell and
benchmark, no-skill baseline $\to$ SkillCorpus condition.
SkillsBench reports pass@1 averaged over three runs
(\%, $n{=}87$ tasks, the $4$ that do not build scored as
failures).  GDPVal reports
the mean LLM-judge reward (3-rep) over all $220$ tasks.
QwenClawBench reports the mean hybrid
score (automated checks and an LLM judge), 3-rep mean.  Rewards are
on $[0,1]$, shown $\times 100$.  Per-cell gains are visualised in
Figure~\ref{fig:main_heatmap}.
The bottom row gives the pooled $\Delta$ across the four cells,
with a task-clustered standard error over $n{=}87/220/100$ tasks
and two-sided $z{=}\Delta/\mathrm{SE}$.}
\label{tab:main-eff}
\end{table*}

\paragraph{Benchmarks, harnesses, and backbones.}
We evaluate on three third-party agent benchmarks, chosen
because frontier LLMs do not saturate them and their domain
coverage is complementary.  SkillsBench \citep{SkillsBench2026}
contributes $87$ deterministic-verifier tasks across $8$
domains and is the closest direct analog to our setting.
GDPVal \citep{OpenAIGDPVal2025} contributes $220$ real-world
economic tasks that span legal, financial, design, and writing
work.  QwenClawBench \citep{QwenClawBench2026} adds $100$ agent
tasks across $8$ domains drawn from a real-user task
distribution.
In total the evaluation covers $407$ tasks spanning $26$
domain labels (Appendix~\ref{app:coverage},
Figure~\ref{fig:domain_coverage}).
We use two
open-source \texttt{SKILL.md}-conformant agent harnesses,
OpenClaw~\citep{OpenClaw2025} and Raven~\citep{Raven2026},
developed independently, paired with two open backbones that
span the lightweight-to-large range, Qwen3.5-27B (Q-27B) and
Qwen3.5-397B-A17B-GPTQ-Int4 (Q-397B), yielding four
(harness, backbone) cells.  A frontier-model robustness check on
Claude~Opus~4.7 verifies that the effect
is not specific to the open backbones.

\paragraph{Configurations, inference, and metrics.}
Each cell is evaluated under a no-skill baseline and under
SkillCorpus paired with our retrieval-and-selection stack
(\S\ref{sec:retrieval}), whose standalone retrieval performance
is reported in \S\ref{sec:retrieval-perf}.  The per-cell
$\Delta$ therefore measures the end-to-end pipeline effect.  All
runs use each backbone's API-default
inference configuration.  The Qwen models run greedy decoding
with their default context and tool-use template, and Opus~4.7
runs at its default thinking effort.  \ifarxiv
Both conditions in a cell
share the same setting, so the reported within-cell $\Delta$ is
unaffected by inference effort and remains valid even though our
vendor-default absolute baselines are not directly comparable to
published leaderboard numbers obtained under other
configurations.
\else
Both conditions in a cell share the same setting, so the within-cell
$\Delta$ stays valid even though our vendor-default absolute
baselines are not comparable to published leaderboard numbers.
\fi  For SkillsBench we report the binary
pass rate over all $87$ tasks, scoring the $4$ that
do not build in our environment as failures in both
conditions.  For GDPVal and QwenClawBench we report the mean
continuous reward, both on $[0,1]$ with $\Delta$ in percentage
points.  GDPVal is scored by a GPT-4o judge, from a different
model family than our Qwen and Opus backbones, and QwenClawBench
by its official grading configuration combining automated checks
with an LLM judge \citep{QwenClawBench2026}.  The main grid holds $24$ configurations, the four cells
evaluated on three benchmarks under both conditions, each
averaged over three runs.  With the two-condition single-run
frontier check, this comes to $74$ end-to-end runs.
Appendix~\ref{app:compute} reports the ingest and evaluation
compute.

% =============================================================================
\subsection{Main results}
\label{sec:main-results}

\begin{figure}[!t]
\centering
\includegraphics[width=\columnwidth]{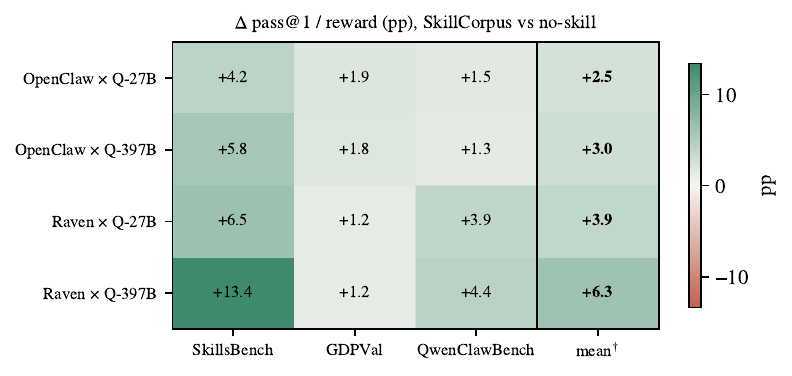}
\caption{Per-cell, per-benchmark $\Delta$ (pp) computed from
the absolute values in Table~\ref{tab:main-eff}.  Green =
positive, red = negative.
$^{\dagger}$Mean over the three benchmarks.}
\label{fig:main_heatmap}
\end{figure}

\ifarxiv
Table~\ref{tab:main-eff} reports the absolute performance of
the no-skill baseline and the SkillCorpus condition (the corpus
with our retrieval-and-selection stack) for each of the four
(harness, backbone) cells on each benchmark.  The per-cell
gains ($\Delta$) and their per-cell means are visualised in
Figure~\ref{fig:main_heatmap}.
\else
Table~\ref{tab:main-eff} reports absolute performance for the
no-skill baseline and SkillCorpus across the four cells and three
benchmarks, with the per-cell gains ($\Delta$) and their means in
Figure~\ref{fig:main_heatmap}.
\fi

% SkillsBench column = get_result_all.py 3-rep mean over jobs_6.3/6.6/6.8,
% 87-denom (4 unbuilt=fail); Δ mean±SE (per cell, n=87):
% oc27 +4.2±2.7, oc397 +5.7±3.1, ec27 +6.5±3.0, ec397 +13.4±3.1;
% pooled +7.47, task-clustered SE 2.35 (z=3.2). (Replaces old +8.4
% 8way_summary headline, 2026-06-16.)
% SE convention = task-clustered: per-task Δ averaged over 4 cells, then
% std/√n over n tasks (NOT pooled cell-task n=4×tasks, which understates SE).
% GDPVal column = FINAL 220 full-set 3-rep mean (0616 regrade,
% analysis_cache/gdpval_percase_full.csv); retry-dedup = keep-original
% (per (task,round) keep __trial_1, drop _retry*; matches oc cells which have
%  no retries; 2026-06-25). Δ: oc27 +1.9, oc397 +1.8, ec27 +1.2, ec397 +1.2;
% pooled +1.51, SE 0.49 (z=3.1). ec base lifted ~2pp vs old pool-all conv.
% (ec27 80.5->82.6, ec397 81.9->84.0) since originals score above retries.
% (regrade 0616 lifted mis-graded skill-arm rows upward, all on skill arm;
%  no-skill arm aggregate unchanged. Confirmed real grader errors.)
% QwenClawBench column = FINAL C (3-round splice, patched gate + reference-usage
% prompt, 0616). Δ: oc27 +1.5, oc397 +1.4,
% ec27 +3.9, ec397 +4.4; pooled +2.79, task-clustered SE 0.70 (z=4.0).
On SkillsBench, which has a
published oracle baseline \citep{SkillsBench2026}, all four
open-backbone cells show positive $\Delta$ averaging
$+7.5$\,pp over three runs, from $+4.2{\pm}2.7$ on the weakest
cell to $+13.4{\pm}3.1$ on the strongest
(Raven~$\times$~Q-397B), with an aligned $+8.0$\,pp on the
Claude~Opus~4.7 frontier check.  In one
run the strongest cell rescues $19$
baseline-failing tasks while breaking $2$ (exact McNemar test,
$p{<}0.001$).  No cell shows a net-negative mean, so at the cell-mean level the
corpus behaves as a no-harm attachment whose upside varies by
cell, even though individual tasks can still regress
(\S\ref{sec:negative}).

\paragraph{Consistent positive effect across benchmarks.}
\ifarxiv
We quantify the effect at the benchmark level by averaging each
task's paired delta across the four cells and computing a
task-clustered standard error, $\mathrm{SE}=\mathrm{std}/\sqrt{n}$
over the $n$ tasks, so that repeated measurements of the same task
across cells are not treated as independent observations.  This is
the more conservative of two reasonable clusterings.  Pooling the
cell--task deltas as independent observations would roughly halve
the standard errors, and the positive effect holds
under either choice.
\else
We quantify the effect at the benchmark level by averaging each
task's paired delta across the four cells and computing a
task-clustered standard error ($\mathrm{SE}=\mathrm{std}/\sqrt{n}$
over the $n$ tasks), so repeated measurements of the same task are
not treated as independent.  This is the more conservative
clustering, and pooling the cell--task deltas would roughly halve
the SEs, but the effect holds either way.
\fi
All
three benchmarks show a consistent positive effect, each
exceeding twice its standard error (bottom row of
Table~\ref{tab:main-eff}).
\ifarxiv
Effect size broadly tracks the headroom a benchmark leaves.
SkillsBench starts near a $10\%$ no-skill baseline and
gains the most, and within it the low-baseline
Industrial~\&~Physical category gains most ($+45.5$\,pp), whereas
GDPVal and QwenClawBench start at $65$--$85\%$ and gain less, where
ceiling effects and judge noise both bound the realisable gain.  Per-task repetition noise is large
on the high-baseline benchmarks, with per-task reward standard
deviations of $20$--$27$\,pp on GDPVal, so most individual cells
are not separately distinguishable from zero.
\else
Effect size broadly tracks the headroom a benchmark leaves.
SkillsBench starts near a $10\%$ no-skill baseline and gains the
most, and within it the low-baseline Industrial~\&~Physical
category gains most ($+45.5$\,pp), whereas GDPVal and QwenClawBench
start at $65$--$85\%$ and gain less, bounded by ceiling effects and
judge noise.  Per-task reward SDs of $20$--$27$\,pp on GDPVal leave
most individual cells indistinguishable from zero.
\fi  \ifarxiv
We do
not observe the negative tail that \citet{SkillsBench2026} reports
for hand-attached skills, nor the failure to improve over a
no-skill baseline reported for uncurated community libraries
\citep{SkillFlow2026,UCSBSkillsWild2026}, a difference consistent with the release gates that
remove the low-quality mass driving them.
\else
We see neither the negative tail \citet{SkillsBench2026} reports for
hand-attached skills nor the no-improvement result for uncurated
libraries \citep{SkillFlow2026,UCSBSkillsWild2026}, consistent with
our release gates removing the low-quality mass driving them.
\fi  Raven outgains OpenClaw on SkillsBench and QwenClawBench
across both backbones.  On GDPVal neither harness leads, as the
gains there are small and individually not significant.  The harness
thus shapes how much of the gain is realised
(\S\ref{sec:discussion}).

% =============================================================================
\subsection{Frontier robustness check}
\label{sec:frontier}
The four main-grid cells span the open capability range
(Q-27B, Q-397B).  To verify that the pipeline's effect is not
specific to this range, we add a single-cell check on
Claude~Opus~4.7 on SkillsBench (OpenClaw harness) under the same
two-condition protocol.  Opus~4.7 reaches a $39.1\%$ no-skill baseline on the $87$-task
set, and adding SkillCorpus shifts pass@1 to $47.1\%$, a $\Delta$
of $+8.0$\,pp.  \ifarxiv
Although the Opus baseline is several times the
open-backbone baselines, its gain is of the same order as the
open-backbone cross-cell mean, indicating that stronger frontier
models have not made an external curated skill pipeline redundant
on this benchmark.  We treat this as a robustness check, and a
broader frontier grid is left for follow-up work.
\else
Although the Opus baseline is several times the open-backbone
baselines, its gain is of the same order as the open-backbone mean,
so stronger frontier models have not made an external curated skill
pipeline redundant here.  A broader frontier grid is left for
follow-up.
\fi

% --------------------------------------------------------- 4.2.Y retrieval-perf
\subsection{Standalone retrieval}
\label{sec:retrieval-perf}
\begin{figure}[!ht]
\centering
\includegraphics[width=\columnwidth]{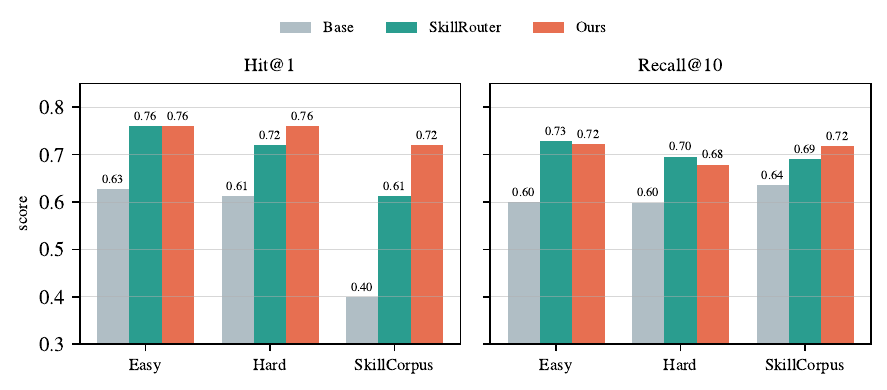}
\caption{Standalone retrieval performance of \textsc{Base},
\textsc{SkillRouter}, and our fine-tuned stack on the three
candidate pools (Hit@1 / Recall@10, exact numbers in
Table~\ref{tab:skill_routing_performance}).}
\label{fig:retrieval_perf}
\end{figure}

End-to-end $\Delta$ entangles corpus quality with retrieval
quality.  We additionally evaluate the recall-and-rerank stack as a
standalone retrieval system on the SkillRouter Easy and Hard
tiers \citep{SkillRouter2026} and on our SkillCorpus pool, a
75-task core SkillsBench set.  Figure~\ref{fig:retrieval_perf}
reports Hit@1 and Recall@10 across the three pools, with full
setup and per-method results in Appendix~\ref{app:retrieval-perf}.

Despite being trained on a different pool, our stack matches
\textsc{SkillRouter} on the Easy tier and exceeds it on the Hard
tier, staying within $1.7$\,pp on Recall@10.  On SkillCorpus it
outperforms \textsc{SkillRouter} by $+10.7$\,pp Hit@1 and
$+2.7$\,pp Recall@10, a margin that partly reflects the matched
train and evaluation distribution and the ground-truth skills
injected into this pool (Appendix~\ref{app:retrieval-perf}).

% =============================================================================
\subsection{Ablation: corpus and retrieval}
\label{sec:ablation}

To gauge how much of the gain comes from corpus curation versus
the fine-tuned retrieval stack, we lesion each in turn on the
strongest cell (Raven~$\times$~Q-397B, SkillsBench).  Replacing
our retrieval stack with off-the-shelf Qwen3 embedding and
reranking models lowers pass@1 from $22.6\%$ to $13.8\%$, and
replacing the filtered corpus with the $\sim$$821{,}000$-file raw
crawl lowers it to $14.9\%$ (Table~\ref{tab:ablation}).  \ifarxiv
Both remain well
above the $9.2\%$ no-skill baseline yet well below the full
pipeline, so curation and retrieval each contribute and neither
alone accounts for the full gain, though the end-to-end $\Delta$
also folds in the LLM rewriter/selector layer that this ablation
does not isolate.  These are single-run numbers
with inter-arm gaps of only a few tasks, so we read them as
directional and leave a repeated ablation to future work.
\else
Both remain well above the $9.2\%$ no-skill baseline yet below the
full pipeline, so curation and retrieval each contribute and neither
alone accounts for the full gain, which also folds in the LLM
rewriter/selector layer this ablation does not isolate.  These
single-run numbers have inter-arm gaps of only a few tasks, so we
read them as directional and leave a repeated ablation to future
work.
\fi

\begin{table}[!ht]
\centering\small
\setlength{\tabcolsep}{6pt}
\begin{tabular}{@{}llcc@{}}
\toprule
Corpus & Retrieval stack & pass@1 & $\Delta$ \\
\midrule
\multicolumn{2}{@{}l}{No-skill baseline} & $9.2$ & --- \\
\midrule
Filtered  & Fine-tuned (ours)   & $\mathbf{22.6}$ & $\mathbf{+13.4}$ \\
Filtered  & Off-the-shelf Qwen3  & $13.8$          & $+4.6$ \\
Raw crawl & Fine-tuned (ours)   & $14.9$          & $+5.7$ \\
\bottomrule
\end{tabular}
\caption{Single-run ablation on the strongest cell
(Raven~$\times$~Q-397B, SkillsBench, pass@1 in $\%$ over $87$
tasks).  The full pipeline pairs the filtered corpus with our
fine-tuned stack, and each lesion swaps exactly one component.
Replacing either the corpus (filtered~$\to$~raw) or the retrieval
stack (ours~$\to$~off-the-shelf) drops the single-run pass@1 to
roughly $14\%$.}
\label{tab:ablation}
\end{table}

% =============================================================================
\subsection{Discussion: where the pipeline helps, where it does not}
\label{sec:discussion}

The per-skill quality score and 16-class taxonomy play supporting
roles rather than driving the gains directly.\label{sec:claim-a}
The taxonomy steers selection benchmark-adaptively, its post-gate
class distribution concentrating on each benchmark's task-relevant
classes (Appendix~\ref{app:class-bench-align}).  The composite
quality score serves deduplication tie-breaks and retrieval-time
ranking rather than per-task prediction, and the active set is
gated by safety and licence, not by a score threshold, with
only a weak safety-facet signal (Appendix~\ref{app:quality-detail}).
\ifarxiv\else
These labels come from a single LLM judge scoring \texttt{SKILL.md}
text rather than from sandboxed execution or human annotation, a
deliberate trade-off at crawl scale that leaves judge bias in the
released set.
\fi

\paragraph{When the pipeline helps.}
Across cells, the per-cell $\Delta$ is largest when the benchmark
contains procedural sub-tasks beyond the
model's pretraining coverage and the corpus covers the task's
domain.  SkillsBench and the QwenClawBench
\textsc{DevOps-Infra}/\textsc{Testing}/\textsc{Meta} subset
(Figure~\ref{fig:domain_coverage}) satisfy both and show the
largest cell-level gains.

\paragraph{Why the harness matters.}
\ifarxiv
Both harnesses receive identical pre-computed skill selections,
and traces show that both ingest them, reading and referencing
the injected content.  The difference is what happens afterwards.
Raven realises a far larger gain than OpenClaw on SkillsBench
($+13.4$ versus $+5.8$\,pp on Q-397B), although both receive the
same skills.  The gain also grows with backbone capability, though entangled
with the harness.  Raven's $+6.5$ at Q-27B rises to $+13.4$ at
Q-397B, and the frontier Opus check adds $+8.0$\,pp on OpenClaw, so
how much of a skill is realised depends on both the harness and the
backbone.
What follows is a qualitative case analysis of the
two Q-397B cells, not an established mechanism.  On the tasks that
Raven passes and OpenClaw fails, trace inspection suggests OpenClaw
consistently stops after the reasoning phase, writing scripts it
never executes and ending without running the verifier, whereas
Raven completes an execute--verify--fix loop.  This is consistent
with skill content converting into verifier passes mainly through
such a loop, so that skill utility appears to be a joint property
of the corpus and the harness, in line with the harness-utilisation
differences \citet{SkillsBench2026} reports across commercial
harnesses.  On this reading, OpenClaw's weaker gains on SkillsBench
are a harness property rather than a corpus one.  The same
execution gap carries no penalty on GDPVal, whose LLM-judged
writing and document tasks reward output quality rather than a
verifier-passing loop, and there the harness gap closes.
\else
Both harnesses receive identical skill selections, and trace
inspection confirms both ingest them, so the difference lies in what
they do next.  Raven realises a far larger SkillsBench gain than
OpenClaw ($+13.4$ versus $+5.8$\,pp on Q-397B), and it grows with
backbone capability (Raven $+6.5$ at Q-27B rising to $+13.4$ at
Q-397B, and the frontier Opus check adds $+8.0$\,pp on OpenClaw), so
realised skill utility depends on both harness and backbone.  As a
qualitative reading of the two Q-397B cells, not an established
mechanism, trace inspection suggests OpenClaw often stops after the
reasoning phase, writing scripts it never executes, whereas Raven
completes an execute--verify--fix loop.  Skill content thus appears
to convert into verifier passes mainly through such a loop, so
OpenClaw's weaker SkillsBench gains read as a harness property rather
than a corpus one, in line with the harness-utilisation differences
\citet{SkillsBench2026} reports.  The same execution gap carries no
penalty on GDPVal, whose LLM-judged tasks reward output quality
rather than a verifier-passing loop, so there the harness gap closes.
\fi

\begin{figure}[!ht]
\centering
\includegraphics[width=\ifarxiv0.82\else0.5\fi\columnwidth]{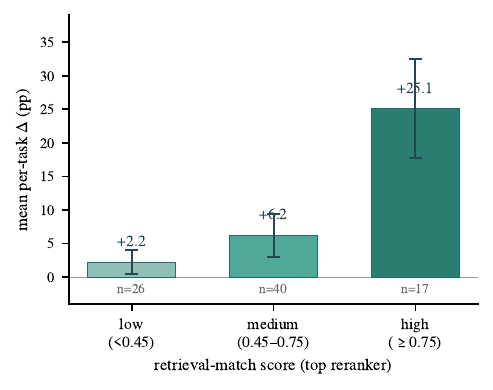}
\caption{Mean per-task SkillsBench gain by retrieval-match bin
(top reranker score of the best-matched skill, a proxy for corpus
coverage), pooled over the four cells, with $\pm$SE error bars.
The score is uncorrelated with the no-skill baseline, so the bins
are not a headroom artefact.  The per-task correlation and raw
scatter are in Appendix~\ref{app:coverage}.}
\label{fig:coverage-binned}
\end{figure}

\paragraph{Coverage shapes the gain.}
\label{sec:negative}
The gains track how well the corpus covers each task.  As a proxy
we take the relevance of the best-retrieved skill (its top reranker
score), which reflects both whether the corpus holds a matching
skill and whether retrieval surfaces it.  Binning tasks by this
retrieval-match score, mean $\Delta$ climbs from $+2.2$\,pp to
$+25.1$\,pp from the lowest to the highest bin
(Figure~\ref{fig:coverage-binned}, pooled over the four cells).
Per task the score correlates with $\Delta$ at Pearson
$r{=}0.31$--$0.40$ across all four cells (all $|t|{>}2.9$, $n{=}83$),
and since it is essentially uncorrelated with the no-skill baseline
the association survives controlling for task difficulty (partial
$r{=}0.34$--$0.40$; Appendix~\ref{app:coverage}).  At the 8-category granularity the gains stay
non-negative throughout, with two thin-coverage categories on the
Raven$\times$Q-397B cell at $\Delta{\approx}0$ (Mathematics,
$N{=}4$; Finance~\&~Economics, $N{=}9$).  Where the corpus lacks
skills for a category, retrieval cannot manufacture them, so the
gain floors at zero rather than turning negative.  Closing these
gaps is a supply problem rather than a retrieval one, calling for
generating or self-evolving the skills the crawl does not yet
contain rather than better matching over the current corpus.  The
rare per-task breaks (two on this cell) stem from procedure--task
mismatch rather than missing coverage, and
Appendix~\ref{app:case-study} dissects one
(\texttt{exceltable-in-ppt}).

% =============================================================================
% §5 Conclusion (~0.3 page)
% =============================================================================
\section{Conclusion}
\label{sec:conclusion}

SkillCorpus curates $\sim$$821{,}000$ crawled community skill files
into a corpus of $96{,}401$ skills, scored along utility,
robustness, and safety and matched to tasks by a fine-tuned
retrieval-and-selection stack.  Across our cross-benchmark,
cross-harness, multi-backbone evaluation
(\S\ref{sec:experiments}), integrating the corpus yields
consistent positive gains on all three benchmarks.  They
are large on SkillsBench and modest on GDPVal and QwenClawBench,
strongest on the Raven harness, with per-domain effects that
remain heterogeneous.  Where coverage is thin, growing the corpus
by skill generation is the natural next step.  The 16-class
taxonomy steers benchmark-aware retrieval.  The composite quality score
serves curation and ranking rather than per-task prediction, and
its safety facet is the one component that both gates unsafe
skills and carries a suggestive per-task signal.  
% We intend SkillCorpus to serve as both an open resource and a reusable framework for community skill ecosystems.
We release SkillCorpus as an open, reusable framework for community skill ecosystems; to our knowledge, it is the first end-to-end account of when a curated community corpus improves real agent tasks, and where it does not.
\ifarxiv\else
Our evaluation prioritises breadth over depth, so on the
high-baseline LLM-judged benchmarks the positive effect is
established at the pooled level rather than cell by cell.  The
English-dominant corpus, the two-harness scope, and the 2026-Q2
snapshot bound external validity, leaving broader languages,
harnesses, and human-validated quality labels to future work.
\fi

\section*{Limitations}
\label{sec:limitations}
% =============================================================================
% Limitations (mandatory unnumbered for *ACL since 2022; not counted in main pages)
% =============================================================================

\paragraph{Breadth-over-depth evaluation design.}
The main grid prioritises breadth across twelve
$(\text{harness}, \text{backbone}, \text{benchmark})$
combinations plus a frontier robustness check.  Depth is
uneven by design.  All three benchmarks are run with three
repeated evaluations per cell, and we report task-clustered
standard errors over the per-task paired deltas
(\S\ref{sec:main-results}).  On the high-baseline LLM-judged
benchmarks (GDPVal, QwenClawBench) per-task judge noise exceeds
the effect size, so the positive effect is established at the
pooled benchmark level rather than cell by cell.  SkillsBench
additionally admits paired per-task statistics (exact McNemar on
a representative run) and the logistic safety-facet regression
(Appendix~\ref{app:quality-detail}).  A larger
number of repetitions and seeds per cell would tighten the
per-cell estimates that the pooled analysis currently leans on.

\paragraph{Text-based skill scoring.}
The 3-facet quality framework and 19-flag taxonomy are evaluated
by an LLM judge against the \texttt{SKILL.md} text content rather
than by sandboxed execution.  This is a deliberate scale
trade-off.  Judge-time text scoring is feasible at the
$\sim$$821{,}000$-file crawl scale, whereas runtime sandboxing
would require per-skill compatible
execution environments not generally available for
community-contributed code.  Safety hard-gates and licence
filtering at Stage~5 catch the most severe content-level risks
at ingest time.  Complementary runtime verification at
retrieval-time or agent-execution-time is a principled extension
for future work.  A second consequence of the scale trade-off is
that the quality, safety, and classification labels driving the
release decision come from a single LLM judge and are not
validated against human gold annotations or cross-checked against
an alternative judge model.  Judge bias therefore propagates into
the released active set.  The facet-separation analysis is an internal-consistency
check, not a
construct-validity study, and human-annotation and
judge-robustness studies are left to future work.

\paragraph{Benchmark and harness coverage.}
The evaluation spans three publicly available benchmarks
(SkillsBench, GDPVal, QwenClawBench) chosen for
non-saturation on frontier LLMs and for complementary task-type
coverage (deterministic-verifier software tasks, real-world
knowledge work, and a
real-user-distribution agent benchmark), and two
\texttt{SKILL.md}-conformant agent harnesses (OpenClaw, Raven)
covering the most widely adopted skill-loading convention in
the public ecosystem.  Per-benchmark and per-harness
heterogeneity within this coverage is reported transparently
in the main results (\S\ref{sec:negative}) rather than
relegated to this section.
Extending the evaluation to additional task types
(e.g., long-horizon embodied tasks, non-English workflows) and
to harnesses with substantially different tool-use interfaces
or planning architectures would broaden external validity.  The
corpus is also English-dominant, both in its source repositories
and in our evaluation, which bounds claims about non-English
skills.  The
corpus-side findings (Stage~5 active--vs--filtered separation,
class-adaptive retrieval, and the suggestive safety-facet signal)
are expected to transfer modulo such adapter work, while per-cell
$\Delta$ magnitudes would re-anchor against each new harness's
own no-skill baseline.

\paragraph{Snapshot semantics.}
The released active set is a snapshot of the public
\texttt{SKILL.md} ecosystem at crawl time, while the community itself
continues to evolve (new repositories, licence updates, shifting
class distributions).  We do not study temporal churn or
re-curation cycles in this work, but the pipeline (per-stage
reject logs, deterministic Stage~5 cut, deduplication keyed on
content hash and embedding cosine) is constructed to support
periodic re-runs against future ecosystem snapshots.  The active
set released with this paper should be read as a 2026-Q2
snapshot rather than a one-shot artifact.

\section*{Ethical Statement}
% =============================================================================
% Ethics Statement (recommended unnumbered for *ACL; not counted in page limit)
% =============================================================================

\ifarxiv
% ---- Full four-paragraph version (arXiv/preprint) ----
\paragraph{Data provenance and licensing.}
SkillCorpus aggregates publicly available \texttt{SKILL.md}
files from GitHub and community marketplaces that link to
GitHub-hosted sources.  Each released skill carries metadata
including its source URL and content hash.  We do not modify
body content or claim authorship.  Every released skill carries
an OSI-approved permissive licence inferred from the source
repository's GitHub \texttt{spdx\_id} and enforced as a Stage-5
ingest filter (Table~\ref{tab:license}, per-reason exclusion
counts in Table~\ref{tab:license-filter}).  Downstream consumers
redistributing the corpus or its derivatives are advised to
perform their own due-diligence audit of \texttt{scripts/}
content and transitive dependencies.

\paragraph{Safety filtering and dual-use.}
The Stage-5 safety hard-gate (\S\ref{sec:classify}) excludes the
most severe categories from the active release set, and the
soft-flag categories are recorded in each skill's metadata for
downstream per-deployment filtering.  We acknowledge the dual-use
risk that a publicly searchable skill corpus could be queried for
adversarial-style content patterns, a residual risk that the
hard-gate and released soft-flag metadata reduce but do not
eliminate.

\paragraph{Benchmark contamination.}
SkillCorpus is community-contributed and pre-dates our benchmark
selection (\S\ref{sec:setup}).  By design the corpus aims for
broad coverage of the skill ecosystem, so released skills overlap
the domains of our evaluation tasks.  These are general procedural
knowledge rather than task-specific solutions, and the per-task
gate injects at most a few topically relevant skills.  We do not
perform task-specific leakage audits beyond the corpus-level
safety and licence filters described above, and consumers using
verifier-backed benchmarks should re-audit for task-specific
contamination when their evaluation regime is sensitive to it.

\else
% ---- Condensed two-paragraph version (AAAI submission, 7-page limit) ----
\paragraph{Data, licensing, and safety.}
SkillCorpus aggregates publicly available \texttt{SKILL.md} files
from GitHub and community marketplaces, recording each skill's
source URL and content hash without modifying its body or claiming
authorship.  Every released skill carries an OSI-approved permissive
licence inferred from the source repository's \texttt{spdx\_id} and
enforced as a Stage-5 gate (Tables~\ref{tab:license}
and~\ref{tab:license-filter}).  The same stage hard-gates the most
severe safety categories and records soft-flags in metadata for
per-deployment filtering.  We acknowledge the residual dual-use risk
that a searchable skill corpus can be queried for adversarial
content, which the gate and flags reduce but do not eliminate.
Downstream consumers redistributing the corpus should audit
\texttt{scripts/} content and transitive dependencies.

\paragraph{Benchmark contamination.}
The corpus is community-contributed and pre-dates our benchmark
selection (\S\ref{sec:setup}).  Its broad coverage overlaps the
domains of our tasks, but the released skills are general procedural
knowledge rather than task-specific solutions, and the per-task gate
injects at most a few topically relevant skills.  We do not perform
task-specific leakage audits beyond the corpus-level filters above,
and consumers using verifier-backed benchmarks should re-audit when
their regime is sensitive to it.
\fi

% =============================================================================
% References — aaai2027.sty already sets \bibliographystyle{aaai2027},
% so no \bibliographystyle command is placed here (per AAAI instructions).
% =============================================================================
\bibliography{refs}

@article{Yao2023ReAct,
  title = {{ReAct}: Synergizing Reasoning and Acting in Language Models},
  author = {Yao, Shunyu and Zhao, Jeffrey and Yu, Dian and Du, Nan and
            Shafran, Izhak and Narasimhan, Karthik and Cao, Yuan},
  journal = {ICLR},
  year = {2023},
  url = {https://arxiv.org/abs/2210.03629}
}

@article{Shinn2023Reflexion,
  title = {Reflexion: Language Agents with Verbal Reinforcement Learning},
  author = {Shinn, Noah and Cassano, Federico and Berman, Edward and
            Gopinath, Ashwin and Narasimhan, Karthik and Yao, Shunyu},
  journal = {NeurIPS},
  year = {2023},
  url = {https://arxiv.org/abs/2303.11366}
}

@inproceedings{Park2023Generative,
  title = {Generative Agents: Interactive Simulacra of Human Behavior},
  author = {Park, Joon Sung and O'Brien, Joseph C. and Cai, Carrie Jun and
            Morris, Meredith Ringel and Liang, Percy and Bernstein,
            Michael S.},
  booktitle = {UIST},
  year = {2023},
  url = {https://arxiv.org/abs/2304.03442}
}

@article{Wang2023Voyager,
  title = {Voyager: An Open-Ended Embodied Agent with Large Language Models},
  author = {Wang, Guanzhi and Xie, Yuqi and Jiang, Yunfan and Mandlekar, Ajay
            and Xiao, Chaowei and Zhu, Yuke and Fan, Linxi and Anandkumar, Anima},
  journal = {Transactions on Machine Learning Research},
  year = {2024},
  url = {https://arxiv.org/abs/2305.16291}
}

@inproceedings{Schick2023Toolformer,
  title = {Toolformer: Language Models Can Teach Themselves to Use Tools},
  author = {Schick, Timo and Dwivedi-Yu, Jane and Dess{\`\i}, Roberto and
            Raileanu, Roberta and Lomeli, Maria and Hambro, Eric and
            Zettlemoyer, Luke and Cancedda, Nicola and Scialom, Thomas},
  booktitle = {NeurIPS},
  year = {2023},
  url = {https://arxiv.org/abs/2302.04761}
}

@inproceedings{Qin2024ToolLLM,
  title = {ToolLLM: Facilitating Large Language Models to Master
           16000+ Real-world APIs},
  author = {Qin, Yujia and Liang, Shihao and Ye, Yining and Zhu, Kunlun
            and Yan, Lan and Lu, Yaxi and Lin, Yankai and Cong, Xin and Tang,
            Xiangru and Qian, Bill and others},
  booktitle = {ICLR},
  year = {2024},
  url = {https://arxiv.org/abs/2307.16789}
}

@article{Patil2023Gorilla,
  title = {Gorilla: Large Language Model Connected with Massive {APIs}},
  author = {Patil, Shishir G. and Zhang, Tianjun and Wang, Xin and
            Gonzalez, Joseph E.},
  journal = {arXiv preprint arXiv:2305.15334},
  year = {2023},
  url = {https://arxiv.org/abs/2305.15334}
}

@article{Hou2025MCP,
  title = {Model Context Protocol ({MCP}): Landscape, Security Threats,
           and Future Research Directions},
  author = {Hou, Xinyi and Zhao, Yanjie and Wang, Shenao and Wang, Haoyu},
  journal = {ACM Transactions on Software Engineering and Methodology},
  year = {2025},
  doi = {10.1145/3796519},
  note = {arXiv:2503.23278},
  url = {https://arxiv.org/abs/2503.23278}
}

@article{MCPCorpus2025,
  title = {A Large-Scale Evolvable Dataset for Model Context Protocol
           Ecosystem and Security Analysis},
  author = {Lin, Zhiwei and Ruan, Bonan and Liu, Jiahao and Zhao, Weibo},
  journal = {arXiv preprint arXiv:2506.23474},
  year = {2025},
  url = {https://arxiv.org/abs/2506.23474}
}

@article{Zhang2024Memento,
  title = {{Memento-Skills}: Let Agents Design Agents},
  author = {Zhou, Huichi and Guo, Siyuan and Liu, Anjie and Yu, Zhongwei
            and Gong, Ziqin and Zhao, Bowen and Chen, Zhixun and Zhang,
            Menglong and Chen, Yihang and Li, Jinsong and others},
  journal = {arXiv preprint arXiv:2603.18743},
  year = {2026},
  url = {https://arxiv.org/abs/2603.18743}
}

@article{AEvolve2025,
  title = {Position: Agentic Evolution is the Path to Evolving {LLM}s},
  author = {Lin, Minhua and Lu, Hanqing and Shi, Zhan and He, Bing and
            Mao, Rui and Zhang, Zhiwei and Wu, Zongyu and Tang,
            Xianfeng and Liu, Hui and Dai, Zhenwei and Zhang, Xiang and
            Wang, Suhang and Dumoulin, Benoit and Pei, Jian},
  journal = {arXiv preprint arXiv:2602.00359},
  year = {2026},
  url = {https://arxiv.org/abs/2602.00359}
}

@misc{AnthropicSkillCreator2025,
  title = {Anthropic Skill Creator: A Rubric for Agent Skill Quality},
  author = {{Anthropic}},
  year = {2025},
  note = {anthropic.com/blog, 2025.}
}

@article{SkillNet2026,
  title = {{SkillNet}: Create, Evaluate, and Connect {AI} Skills},
  author = {Liang, Yuan and Zhong, Ruobin and Xu, Haoming and Jiang, Chen
            and Zhong, Yi and Fang, Runnan and Gu, Jia-Chen and Deng,
            Shumin and Yao, Yunzhi and Wang, Mengru and Qiao, Shuofei
            and Xu, Xin and Wu, Tongtong and Wang, Kun and Liu, Yang and
            others},
  journal = {arXiv preprint arXiv:2603.04448},
  year = {2026},
  url = {https://arxiv.org/abs/2603.04448}
}

@article{SkillsBench2026,
  title = {{SkillsBench}: Benchmarking How Well Agent Skills Work Across
           Diverse Tasks},
  author = {Li, Xiangyi and Chen, Wenbo and Liu, Yimin and Zheng, Shenghan
            and Chen, Xiaokun and He, Yifeng and Li, Yubo and You, Bingran
            and Shen, Haotian and Sun, Jiankai and others},
  journal = {arXiv preprint arXiv:2602.12670},
  year = {2026},
  url = {https://arxiv.org/abs/2602.12670}
}

@article{SkillFlow2026,
  title = {{SkillFlow}: Scalable and Efficient Agent Skill Retrieval System},
  author = {Li, Fangzhou and Tagkopoulos, Pagkratios and Tagkopoulos, Ilias},
  journal = {arXiv preprint arXiv:2504.06188},
  year = {2025},
  url = {https://arxiv.org/abs/2504.06188}
}

@article{UCSBSkillsWild2026,
  title = {How Well Do Agentic Skills Work in the Wild: Benchmarking {LLM}
           Skill Usage in Realistic Settings},
  author = {Liu, Yujian and Ji, Jiabao and An, Li and Jaakkola, Tommi and
            Zhang, Yang and Chang, Shiyu},
  journal = {arXiv preprint arXiv:2604.04323},
  year = {2026},
  url = {https://arxiv.org/abs/2604.04323}
}

@article{AgentSkillsDataAnalysis2026,
  title = {Agent Skills: A Data-Driven Analysis of {Claude} Skills for
           Extending Large Language Model Functionality},
  author = {Ling, George and Zhong, Shanshan and Huang, Richard},
  journal = {arXiv preprint arXiv:2602.08004},
  year = {2026},
  url = {https://arxiv.org/abs/2602.08004}
}

@article{ScalingLaws2026,
  title = {The Scaling Laws of Skills in {LLM} Agent Systems},
  author = {Chen, Charles and Yu, Qiming and Gu, Yuhang and Huang, Zhuoye
            and Li, Hanjing and Liu, Hongyu and Liu, Simin and Liu, Jinhao
            and Peng, Dengyun and Wang, Jiangyi and Yan, Zheng and Meng,
            Fanqing and Qin, Ethan and Che, Carl and Hu, Mengkang},
  journal = {arXiv preprint arXiv:2605.16508},
  year = {2026},
  url = {https://arxiv.org/abs/2605.16508}
}

@article{SkillRet2026,
  title = {{SkillRet}: A Large-Scale Benchmark for Skill Retrieval in
           {LLM} Agents},
  author = {Cho, Hongcheol and Kang, Ryangkyung and Kim, Youngeun},
  journal = {arXiv preprint arXiv:2605.05726},
  year = {2026},
  url = {https://arxiv.org/abs/2605.05726}
}

@article{SkillRouter2026,
  title = {{SkillRouter}: Skill Routing for {LLM} Agents at Scale},
  author = {Zheng, YanZhao and Zhang, ZhenTao and Ma, Chao and Yu, YuanQiang
            and Zhu, JiHuai and Wu, Yong and Xu, Tianze and Dong, Baohua and
            Zhu, Hangcheng and Huang, Ruohui and Yu, Gang},
  journal = {arXiv preprint arXiv:2603.22455},
  year = {2026},
  url = {https://arxiv.org/abs/2603.22455}
}

@article{GraphOfSkills2026,
  title = {Graph of Skills: Dependency-Aware Structural Retrieval for
           Massive Agent Skills},
  author = {Liu, Dawei and Li, Zongxia and Du, Hongyang and Wu, Xiyang and
            Gui, Shihang and Kuang, Yongbei and Sun, Lichao},
  journal = {arXiv preprint arXiv:2604.05333},
  year = {2026},
  url = {https://arxiv.org/abs/2604.05333}
}

@article{Merrill2026TerminalBench,
  title = {{Terminal-Bench}: Benchmarking Agents on Hard, Realistic Tasks
           in Command Line Interfaces},
  author = {Merrill, M. A. and Shaw, A. G. and Carlini, N. and Li, B. and
            Raj, H. and Bercovich, I. and Shi, L. and Shin, J. Y. and Walshe,
            T. and Buchanan, E. K. and others},
  journal = {arXiv preprint arXiv:2601.11868},
  year = {2026},
  url = {https://arxiv.org/abs/2601.11868}
}

@article{AgentSkillOS2026,
  title = {Organizing, Orchestrating, and Benchmarking Agent Skills at
           Ecosystem Scale},
  author = {Li, Hao and Mu, Chunjiang and Chen, Jianhao and Ren, Siyue
            and Cui, Zhiyao and Zhang, Yiqun and Bai, Lei and Hu, Shuyue},
  journal = {arXiv preprint arXiv:2603.02176},
  year = {2026},
  url = {https://arxiv.org/abs/2603.02176}
}

@article{SkillRAE2026,
  title = {{SkillRAE}: Agent Skill-Based Context Compilation for
           Retrieval-Augmented Execution},
  author = {Meng, Xiangcheng and Wang, Shu and Fang, Yixiang},
  journal = {arXiv preprint arXiv:2605.10114},
  year = {2026},
  url = {https://arxiv.org/abs/2605.10114}
}

@misc{OpenAIGDPVal2025,
  title = {{GDPval}: Measuring {LLM} Performance on Real-World Economic Tasks},
  author = {{OpenAI}},
  year = {2025},
  note = {openai.com/research, 2025.}
}

@misc{QwenClawBench2026,
  title = {{QwenClawBench}: A Real-User-Distribution Agent Benchmark for
           {OpenClaw}},
  author = {{Qwen Team, Alibaba Group}},
  year = {2026},
  note = {Open-source benchmark, github.com/SKYLENAGE-AI/QwenClawBench,
          released April 2026.},
  url = {https://github.com/SKYLENAGE-AI/QwenClawBench}
}

@misc{AnthropicClaudeCode2025,
  title = {{Claude Code}: An Agentic Coding Tool},
  author = {{Anthropic}},
  year = {2025},
  note = {github.com/anthropics/claude-code, 2025.}
}

@misc{OpenAIClaudeCode2025,
  title = {Codex {CLI}: Lightweight Coding Agent that Runs in Your Terminal},
  author = {{OpenAI}},
  year = {2025},
  note = {github.com/openai/codex, 2025.}
}

@misc{OpenClaw2025,
  author = {Peter Steinberger},
  title  = {{OpenClaw}: An Open-Source Autonomous Agent Framework},
  year   = {2025},
  url    = {https://github.com/openclaw/openclaw},
  note   = {Originally published Nov 2025 as Clawdbot; renamed OpenClaw 2026. Skill-loading via \texttt{SKILL.md} convention.}
}

@misc{Raven2026,
  author = {{Raven Contributors}},
  title  = {{Raven}: An Open-Source Agent Framework},
  year   = {2026},
  url    = {https://github.com/EverMind-AI/Raven},
  note   = {Open-source repository; supports \texttt{SKILL.md} skill-loading convention.}
}

@inproceedings{Lee2022Dedup,
  title     = {Deduplicating Training Data Makes Language Models Better},
  author    = {Lee, Katherine and Ippolito, Daphne and Nystrom, Andrew and Zhang, Chiyuan and Eck, Douglas and Callison-Burch, Chris and Carlini, Nicholas},
  booktitle = {Proceedings of the 60th Annual Meeting of the Association for Computational Linguistics (ACL)},
  year      = {2022},
  note      = {arXiv:2107.06499}
}

@inproceedings{Zhou2023LIMA,
  title     = {{LIMA}: Less Is More for Alignment},
  author    = {Zhou, Chunting and Liu, Pengfei and Xu, Puxin and Iyer, Srini and Sun, Jiao and Mao, Yuning and Ma, Xuezhe and Efrat, Avia and Yu, Ping and Yu, Lili and Zhang, Susan and Ghosh, Gargi and Lewis, Mike and Zettlemoyer, Luke and Levy, Omer},
  booktitle = {Advances in Neural Information Processing Systems (NeurIPS)},
  year      = {2023},
  note      = {arXiv:2305.11206}
}

@inproceedings{Lewis2020RAG,
  title     = {Retrieval-Augmented Generation for Knowledge-Intensive {NLP} Tasks},
  author    = {Lewis, Patrick and Perez, Ethan and Piktus, Aleksandra and Petroni, Fabio and Karpukhin, Vladimir and Goyal, Naman and K{\"u}ttler, Heinrich and Lewis, Mike and Yih, Wen-tau and Rockt{\"a}schel, Tim and Riedel, Sebastian and Kiela, Douwe},
  booktitle = {Advances in Neural Information Processing Systems (NeurIPS)},
  year      = {2020},
  note      = {arXiv:2005.11401}
}

@inproceedings{Zheng2023LLMJudge,
  title     = {Judging {LLM}-as-a-Judge with {MT}-Bench and Chatbot Arena},
  author    = {Zheng, Lianmin and Chiang, Wei-Lin and Sheng, Ying and Zhuang, Siyuan and Wu, Zhanghao and Zhuang, Yonghao and Lin, Zi and Li, Zhuohan and Li, Dacheng and Xing, Eric P. and Zhang, Hao and Gonzalez, Joseph E. and Stoica, Ion},
  booktitle = {Advances in Neural Information Processing Systems (NeurIPS)},
  year      = {2023},
  note      = {arXiv:2306.05685}
}

% =============================================================================
% Appendix
% =============================================================================
\appendix
% =============================================================================
% Appendix (after references, self-contained, unlimited length)
% Organised into three thematic parts mirroring the paper: corpus construction
% (§3), retrieval & evaluation (§4), and compute & reproducibility (artifact).
% =============================================================================

% #############################################################################
\section{Corpus Construction Details}
\label{app:corpus-details}

% =============================================================================
\subsection{Quality Flag Vocabulary}
\label{app:flags}

The 19-flag taxonomy used by the LLM quality judge is structured by
its primary semantic dimension.  Flags marked $^\dagger$ are
hard-gate flags whose firing forces the per-skill quality
score to zero and the skill to be excluded from the released active
set.

\begin{itemize}\setlength\itemsep{2pt}
\item Utility-bound (2): \texttt{marketing\_only},
      \texttt{vague\_purpose}.
\item Robustness-bound (6): \texttt{placeholder},
      \texttt{no\_steps}, \texttt{deprecated\_api},
      \texttt{fabricated\_call}, \texttt{syntax\_error},
      \texttt{inconsistent\_doc\_code}.
\item Safety-bound (11): \texttt{destructive\_no\_confirm},
      \texttt{secret\_leak}, \texttt{network\_exfil},
      \texttt{prompt\_injection}$^\dagger$,
      \texttt{cmd\_injection}$^\dagger$,
      \texttt{unsafe\_exec}$^\dagger$,
      \texttt{auth\_bypass}$^\dagger$,
      \texttt{fact\_poisoning}, \texttt{tos\_violation},
      \texttt{csam\_risk}$^\dagger$, \texttt{bias\_content}.
\end{itemize}

Flag classification follows the LLM judge prompt.  The same flag
may in principle constrain more than one facet (e.g., a
\texttt{placeholder} skill primarily caps the robustness facet but
may also indicate poor utility scope), but for scoring we use the
single-facet assignment shown above.

\paragraph{Hard-gate vs soft-signal rationale.}
The five hard-gate flags
(\texttt{prompt\_injection},
\texttt{cmd\_injection},
\texttt{unsafe\_exec},
\texttt{auth\_bypass},
\texttt{csam\_risk})
share two properties that distinguish them from the 14 soft
signals: (i)~a single firing is sufficient to make the skill
unsafe to ship even in conjunction with downstream sandboxing or
partial-load policies, and (ii)~remediation cannot be deferred to
per-deployment filtering without re-vetting the body content.
Soft signals such as \texttt{secret\_leak} or
\texttt{destructive\_no\_confirm}, by contrast, can be mitigated
by deployment-side wrappers (credential rotation, confirmation
prompts) or by skipping affected skills in a particular
deployment without invalidating the broader artifact.  The
distinction is therefore an integrability boundary, not an
absolute-severity ranking.

% =============================================================================
\subsection{Quality-Score Distribution and Facet Separation}
\label{app:quality-detail}

\begin{figure}[!ht]
\centering
\includegraphics[width=\columnwidth]{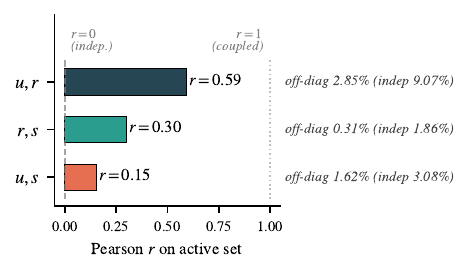}
\caption{Three-facet independence on the active set
($N{=}96{,}401$).  Bar length is the Pearson $r$ for each facet
pair (dashed line at $r{=}0$ marks statistical independence).
Italic annotations report off-diagonal mass
$\Pr(a{\geq}8\wedge b{\leq}4)$ versus the independence baseline.}
\label{fig:facet_pairs}
\end{figure}

Whether the 3-facet prompt produces genuinely separate scores is
subtler than Pearson $r$ alone can answer, because real skills
share latent craftsmanship that elevates all facets
simultaneously, so some positive correlation is expected even
under independent measurement.  For each facet pair we report both
Pearson $r$ and the off-diagonal mass
$\Pr(a{\geq}8 \wedge b{\leq}4)$ against the product-of-marginals
independence baseline, computed on the $96{,}401$-skill active set
and visualised in Figure~\ref{fig:facet_pairs}.  Observed
off-diagonal mass sits below the independence baseline for all
three pairs.  Among the craftsmanship-coupled pair, the prompt
still isolates a $2.85\%$ ``description-good / body-weak'' quadrant
($u{\geq}8 \wedge r{\leq}4$) that a one-dimensional composite score
cannot represent.  Because the subscores are integer $0$--$10$
values subject to soft-flag capping, the Pearson $r$ is an
attenuated estimate of association.

\paragraph{Per-task outcome relationship.}
The composite score does not predict per-task pass rate on any
benchmark (all $|r|<0.10$, $p>0.35$).  Only the safety facet
carries a suggestive per-task signal, a pooled SkillsBench logistic
giving $\hat{\beta}_{\mathrm{safety}}{=}+1.03$ ($z{=}+2.29$,
$p{=}0.022$) that rests on a single grid and would not survive a
multiple-comparison correction, so whether it can serve as a
ranking cue beyond topical relevance merits validation on further
grids.

\paragraph{Score distribution.}
On the $96{,}401$ active skills the composite quality score
(on the $[0,1]$ scale) has mean $0.690$.  The bucket-level
distribution is in Table~\ref{tab:q-dist}.

\begin{table}[!ht]
\centering\small
\setlength{\tabcolsep}{4pt}
\begin{tabular}{@{}lr@{}}
\toprule
Quality bucket          & \% of active \\
\midrule
$q < 0.3$               &  2.8 \\
$q \in [0.3, 0.5)$      &  5.5 \\
$q \in [0.5, 0.7)$      & 32.9 \\
$q \geq 0.7$ (curated)  & 58.8 \\
\bottomrule
\end{tabular}
\caption{Quality-bucket distribution on the active set.}
\label{tab:q-dist}
\end{table}

% =============================================================================
\subsection{Source Prior}
\label{app:source-prior}

Each source $s$ in the corpus is assigned a prior
\begin{align*}
  \mathtt{prior}(s) =\;& 0.95 \cdot \mathtt{track}(s) \\
                       & + 0.05 \cdot \mathtt{lic\_rate}(s) \\
                       & + 0.10 \cdot \mathbb{1}[s \in \mathcal{T}],
\end{align*}
clipped to $[0.30, 1.00]$, where:
\begin{itemize}
\item $\mathtt{track}(s) = \dfrac{n\bar{q}_s + K\mu_0}{n + K}$ is
      the Bayesian-shrunk source quality, with $n$ the number of
      skills from $s$, $\bar{q}_s \in [0,1]$ the source's empirical
      normalised content-quality mean (\S\ref{sec:quality}), $K=10$
      a shrinkage strength in skill-count units, and $\mu_0=0.685$
      the dataset-wide normalised content-quality mean.
\item $\mathtt{lic\_rate}(s) \in [0,1]$ is the fraction of $s$'s
      skills carrying an OSI-permissive licence.
\item $\mathcal{T}$ is a small explicit trust list
      (\texttt{anthropics}, \texttt{antigravity},
      \texttt{karanb192}).  The indicator contributes at most
      $+0.10$ to dampen the influence of hand-curation relative to
      the data-driven track score.
\end{itemize}
$K=10$ was chosen \emph{a priori} from the heuristic "ten observations
is the minimum credible sample".  We did not tune it on downstream
task performance.

% =============================================================================
\subsection{False-Positive Audit of Safety Regex Patterns}
\label{app:fp-audit}

The Stage~5 safety hard-gate applies a single pre-judge malware
regex (\texttt{blocked.malware}).  A set of additional regex patterns
(e.g., suspicious-string heuristics) are retained for audit logging
only.  A stratified 100-skill manual audit indicates that several
audit-only regex patterns have empirical false-positive rates above
$90\%$, which is why they are not used as filtering gates.  Detailed
false-positive counts per regex pattern are provided in the released
ingest report.

% =============================================================================
\subsection{Licence Breakdown and Exclusion Counts}
\label{app:licence}

All $96{,}401$ released skills carry an OSI-approved permissive
licence.  Table~\ref{tab:license} reports the licence composition
of the active set, dominated by MIT ($88.3\%$) and Apache-2.0
($10.5\%$).  Table~\ref{tab:license-filter} itemises the
$3{,}795$ skills excluded at the Stage-5 licence filter by
exclusion reason.  Missing \texttt{LICENSE} files and unreachable
source repositories account for $\sim$$87\%$ of exclusions,
while explicitly incompatible terms (copyleft or
non-commercial/share-alike) account for under $9\%$.

\begin{table}[!ht]
\centering\small
\setlength{\tabcolsep}{3pt}
\begin{tabular}{@{}lr@{}}
\toprule
Licence bucket (all GREEN)                    & \%    \\
\midrule
MIT / MIT-0                                   & 88.3  \\
Apache-2.0                                    & 10.5  \\
Other OSI-permissive$^{\dagger}$              &  1.2  \\
\midrule
Total                                         & 100.0 \\
\bottomrule
\end{tabular}
\caption{Licence breakdown of the released active set
($N{=}96{,}401$).  $^{\dagger}$CC0 / BSD / MPL-2.0 / Unlicense /
ISC / Mulan-PSL-2.0 / WTFPL / 0BSD.}
\label{tab:license}
\end{table}

\begin{table}[!ht]
\centering\small
\setlength{\tabcolsep}{3pt}
\begin{tabular}{@{}lr@{}}
\toprule
Exclusion reason                              & count   \\
\midrule
Source repo has no \texttt{LICENSE} file      & 1{,}756 \\
Source repo unreachable via GitHub API        & 1{,}563 \\
Copyleft (GPL / AGPL / LGPL)                  &   228   \\
Non-commercial or share-alike (CC-NC/SA/ND)   &    86   \\
Custom or unparsed licence string             &   162   \\
\midrule
Total excluded                                & 3{,}795 \\
\bottomrule
\end{tabular}
\caption{Skills excluded at the Stage~5 licence filter to
guarantee commercial redistributability.}
\label{tab:license-filter}
\end{table}

% =============================================================================
\subsection{Crawl Source Inventory}
\label{app:crawl-sources}

The SkillCorpus crawl (\S\ref{sec:pipeline}) is configured by a
single machine-readable registry of $62$ sources.  Each entry
declares one of five ingestion mechanisms, and all mechanisms
feed a shared discovery entry point, so adding a source is a
one-line registry change.  The mechanisms are itemised below
with per-mechanism counts and representative sources.

\begin{table}[!ht]
\centering\scriptsize
\setlength{\tabcolsep}{4pt}
\begin{tabular}{@{}>{\raggedright\arraybackslash}p{0.28\columnwidth}>{\raggedright\arraybackslash}p{0.66\columnwidth}@{}}
\toprule
Mechanism (sources) & Representative sources \\
\midrule
Direct repository clones (38) &
\texttt{anthropics/skills}, \texttt{openclaw/skills},
\texttt{openclaw/clawhub}, \texttt{antigravity}, and the long
tail of repositories tagged \texttt{claude-skills} or
\texttt{agent-skills}. \\
\midrule
Awesome-list README scrapes (19) &
\texttt{VoltAgent/awesome-agent-skills},\newline
\texttt{karanb192/awesome-claude-skills},\newline
\texttt{Chat2AnyLLM/awesome-claude-skills}, and others. \\
\midrule
Marketplace index APIs (2) &
SkillsMP, SkillsDirectory. \\
\midrule
Sitemap crawl (1) &
Skills.sh. \\
\midrule
JSON catalogs (2) &
LobeHub index, SkillManager. \\
\bottomrule
\end{tabular}
\caption{Crawl sources by ingestion mechanism.  The complete
registry ships with the released artifact as a machine-readable
file.}
\label{tab:crawl-sources}
\end{table}

% #############################################################################
\section{Retrieval and Evaluation Details}
\label{app:eval-details}

% =============================================================================
\subsection{Standalone Retrieval Performance}
\label{app:retrieval-perf}

We evaluate the recall-and-rerank stack as a standalone retrieval
system on three candidate pools using the SkillRouter evaluation
protocol \citep{SkillRouter2026}: the SkillRouter Easy tier
(standard large-pool retrieval), the SkillRouter Hard tier (with
adversarial distractors), and our own SkillCorpus pool (with
the ground-truth SkillsBench skills explicitly injected at
evaluation time but excluded from training and from the
downstream agent).  We compare against two baselines:
\textsc{Base} (off-the-shelf Qwen3-Emb-0.6B and Qwen3-Rank-0.6B)
and \textsc{SkillRouter} \citep{SkillRouter2026} using its
released models.  Performance is measured with Hit@1 and
Recall@10 on the $75$ core SkillsBench tasks (excluding $12$
generic-only cases per the SkillRouter protocol).

\begin{table}[!ht]
\centering\small
\setlength{\tabcolsep}{4pt}
\begin{tabular}{@{}llccc@{}}
\toprule
Method & Metric & Easy & Hard & SkillCorpus \\
\midrule
\multirow{2}{*}{\textsc{Base}}
                             & Hit@1     & $0.627$ & $0.613$ & $0.400$ \\
                             & Recall@10 & $0.600$ & $0.599$ & $0.635$ \\
\midrule
\multirow{2}{*}{\textsc{SkillRouter}}
                             & Hit@1     & $\mathbf{0.760}$ & $0.720$ & $0.613$ \\
                             & Recall@10 & $\mathbf{0.729}$ & $\mathbf{0.696}$ & $0.691$ \\
\midrule
\multirow{2}{*}{\textsc{Ours}}
                             & Hit@1     & $\mathbf{0.760}$ & $\mathbf{0.760}$ & $\mathbf{0.720}$ \\
                             & Recall@10 & $0.722$ & $0.679$ & $\mathbf{0.718}$ \\
\bottomrule
\end{tabular}
\caption{Standalone retrieval performance (Hit@1 and Recall@10
on the $75$-task core SkillsBench evaluation set).}
\label{tab:skill_routing_performance}
\end{table}

The Easy / Hard performance shows cross-pool generalisability,
since our stack is competitive on tiers it was not trained on.  The
SkillCorpus performance reflects the expected advantage of
train-time and evaluation-time distributions matching.  The
gap between Easy and SkillCorpus for both \textsc{SkillRouter}
and \textsc{Ours} ($-14.7$ and $-4.0$\,pp on Hit@1 respectively)
indicates that SkillCorpus is intrinsically harder than the Easy
tier, consistent with its larger size and more diverse
distractor mass.

% =============================================================================
\subsection{Cross-Benchmark Domain Coverage}
\label{app:coverage}

\begin{figure*}[!t]
\centering
\includegraphics[width=\textwidth]{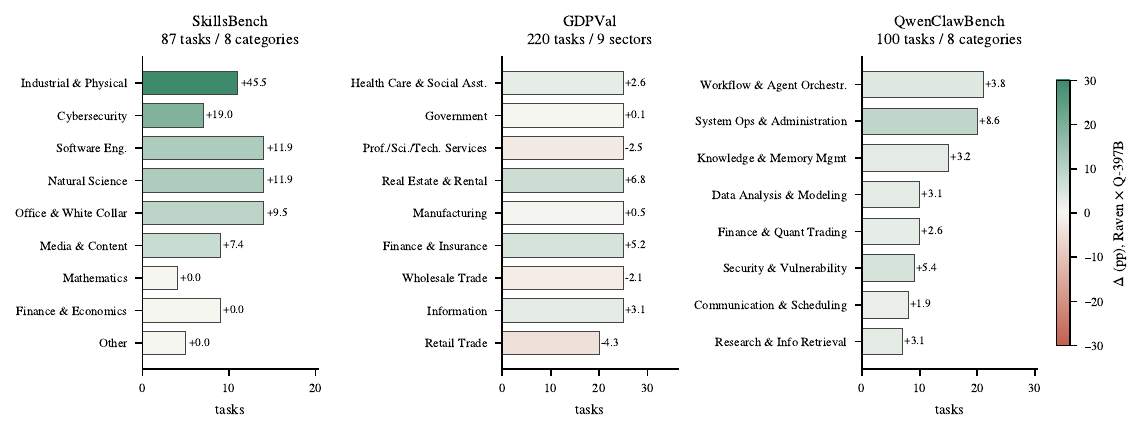}
\caption{Cross-benchmark domain coverage of the evaluation, each
panel in its benchmark's native taxonomy.}
\label{fig:domain_coverage}
\end{figure*}

\noindent Figure~\ref{fig:domain_coverage} documents the breadth
of task domains the corpus is exercised against.  It plots the
$407$ evaluated tasks over $26$ domain labels in each benchmark's
native taxonomy ($8$ SkillsBench categories plus a residual Other
bin, $9$ GDPVal sectors, and $8$ QwenClawBench categories), with
bar length giving the task count and bar colour the per-domain
$\Delta$ on the strongest cell (Raven~$\times$~Q-397B).  Those
per-domain $\Delta$ values are coarse and noisy at $N\,{\le}\,25$
per cell, so they mark only where coverage is dense or thin and
are not per-domain effect claims, which the main text makes at the
pooled level (\S\ref{sec:main-results}).

\begin{figure}[!ht]
\centering
\includegraphics[width=0.86\columnwidth]{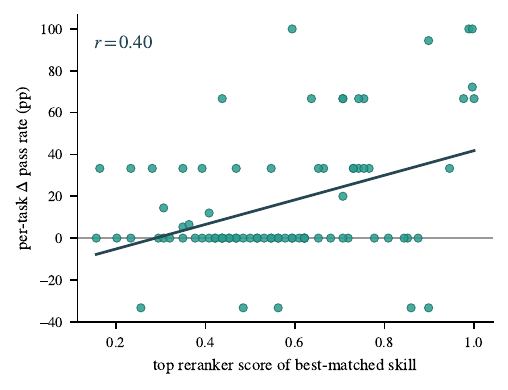}
\caption{Per-task SkillsBench gain versus corpus coverage
(top reranker score of the best-matched skill),
Raven~$\times$~Q-397B ($n{=}83$, $r{=}0.40$).}
\label{fig:coverage}
\end{figure}

At the task level, the relevance of the best-retrieved skill
predicts the gain more directly than these coarse domain bins.
Beyond the binned summary in the main text
(Figure~\ref{fig:coverage-binned}), Figure~\ref{fig:coverage} plots
each SkillsBench task's $\Delta$ against the top reranker score of
its best-matched skill, a proxy that reflects both whether the
corpus holds a matching skill and whether retrieval surfaces it.
The positive slope holds across all four cells (Pearson
$r{=}0.31$--$0.40$, all $|t|{>}2.9$, $n{=}83$), the panel showing the
strongest cell.  The score is essentially uncorrelated with the
no-skill baseline ($|r|<0.05$), so partialling out task difficulty
leaves the association unchanged (partial $r{=}0.34$--$0.40$).

% =============================================================================
\subsection{Class-Benchmark Alignment}
\label{app:class-bench-align}

\noindent Table~\ref{tab:class-bench-align} reports the
per-benchmark class over-selection underlying the class-adaptive
selection result of \S\ref{sec:claim-a}.  For each benchmark it
lists the skill classes most over-represented in the post-gate
selected pool relative to their corpus-wide frequency, showing
that each concentrates the gate on a different, task-appropriate
set of classes.  All three reject the null of the same class
distribution as the corpus (Pearson $\chi^2$ goodness-of-fit,
$p<10^{-10}$).

\begin{table}[!ht]
\centering\small
\setlength{\tabcolsep}{4pt}
\begin{tabular}{@{}lp{0.62\columnwidth}@{}}
\toprule
Benchmark & Over-selected classes (post-gate \% vs corpus \%) \\
\midrule
SkillsBench   & \textsc{Data} (38 vs 14), \textsc{Doc-Proc} (13 vs 3), \textsc{Security} (6 vs 4) \\
GDPVal        & \textsc{Writing} (31 vs 8), \textsc{Data} (29 vs 15), \textsc{Workflow} (14 vs 5) \\
QwenClawBench & \textsc{DevOps-Infra} (13 vs 8), \textsc{Meta} (11 vs 4), \textsc{Testing} (10 vs 6) \\
\bottomrule
\end{tabular}
\caption{Per-benchmark 16-class over-selection relative to the
corpus-wide baseline.}
\label{tab:class-bench-align}
\end{table}

% =============================================================================
\subsection{Case Studies: a Rescue and a Break}
\label{app:case-study}

We illustrate the end-to-end pipeline with two worked examples,
one task that skills rescue and one of the rare tasks that
skills break.  The rescue is
\texttt{manufacturing-\allowbreak codebook-\allowbreak normalization}, a SkillsBench
industrial-physical-systems task that the frontier check of
\S\ref{sec:frontier} (Claude~Opus~4.7 on the OpenClaw harness)
fails without skills and passes with them.

\paragraph{Task.}
Normalise free-form defect-reason texts from manufacturing test
logs (typos, abbreviations, mixed-language fragments) into a
standardised codebook: segment each raw text, assign a valid
code per segment subject to per-station code validity, and emit
calibrated confidence scores that distinguish \textsc{Unknown}
from low-evidence known predictions.

\paragraph{Retrieved skills.}
The gate (\S\ref{sec:retrieval}) selects two skills from the
corpus: a domain skill,
\texttt{manufacturing-\allowbreak failure-\allowbreak reason-\allowbreak codebook-\allowbreak normalization},
and a generic utility, \texttt{fuzzy-match}.  The domain skill's
body specifies the procedure rather than the topic:

\begin{quote}\small
``If \texttt{entry.stations} is not \texttt{None}, the predicted
code should only be considered valid when the record station
matches [\ldots] otherwise the code should be rejected.
[\ldots] consider multiple evidence sources [\ldots] (overlap or
fuzzy similarity between \texttt{span\_text} and codebook text
[\ldots], station compatibility, \texttt{fail\_code} alignment,
\texttt{test\_item} alignment, and conflict cues), and normalise
the score to a stable range $[0.0, 1.0]$.''
\end{quote}

\paragraph{Without skills.}
The agent writes a plausible solution built on a hard-coded
phrase dictionary with station filtering, but no multi-evidence
scoring and no confidence calibration.  It passes $15$ of $16$
verifier tests and fails the semantic-alignment check.  Only
$66.2\%$ of its non-\textsc{Unknown} predictions carry
non-trivial lexical evidence against the codebook (the verifier
requires $\geq 70\%$), i.e.\ it confidently emits codes the text
does not support.  Binary task reward: $0$.

\paragraph{With skills.}
The agent implements the skill's procedure directly:
multi-evidence weighted scoring, station-scope penalties, a
deterministic tie-break for near-tied candidates, and
calibrated confidence with an explicit \textsc{Unknown} band.
All $16$ tests pass, with $19\%$ less wall-clock time and
$20\%$ fewer total tokens than the failing no-skill run.

\paragraph{The break: \texttt{exceltable-in-ppt}.}
The complementary failure mode appears on one of the two tasks
the strongest open cell (Raven~$\times$~Q-397B) breaks
(no-skill passes, with-skill fails).  The task is to update an
exchange rate inside an Excel table embedded in a
PowerPoint slide, preserving its formulas.  Without skills the
agent improvises a working path through the four-layer
nesting, from the PPTX part through the OLE blob and its
embedded ZIP down to the sheet XML, patches the cell, and
re-packs the blob ($8/8$ verifier
tests pass).  With skills, the gate injects two topically
relevant Office skills whose recipes all operate on
top-level files
(\texttt{load\_workbook("file.xlsx")},
\texttt{Presentation("file.pptx")}).  The agent follows the
recipes, hits an API wall
(\texttt{property 'blob' of '\_OleFormat' has no setter}), and
never replaces the embedded object.  The verifier reads
\texttt{nan} where the table should be ($6/8$).  The skills are
on-topic, but their procedures do not extend to this task's
structure, and having a recipe displaced the improvisation that
succeeds without one.

\paragraph{Takeaway.}
The rescue and the break are two sides of the same property.
Skills transfer procedures, not topics.  When the
injected procedure matches the task's structure, as with the
multi-evidence scoring formulas and tie-breaks above, it
supplies exactly what the model does not reliably improvise.
When it does not, as with flat-file recipes against an embedded
object, following it can displace an improvisation that would
have succeeded.  This
is also one concrete reason composite quality scores cannot
predict per-task outcomes (\S\ref{sec:claim-a}).  The skills in
both cases are well-crafted, and what differs is
procedure--task fit.

% #############################################################################
\section{Compute and Reproducibility}
\label{app:compute-repro}

% =============================================================================
\subsection{Per-Stage Compute}
\label{app:compute}

The ingest pipeline was served on an internal GPU cluster using
Qwen3.5-397B-A17B-GPTQ-Int4 as the LLM judge for quality
scoring, classification, and embedding-based dedup.  This judge
is an open-weight model, publicly available on Hugging Face and
via OpenRouter, so the quality, safety, and classification labels
can be re-derived independently.  The internal cluster refers only
to where we ran it.  The compute
reported below is the cumulative total over the crawl that
produces the released active set.

\paragraph{Ingest compute.}
The ingest pipeline issued approximately $269{,}000$ LLM-judge
calls in total: $101{,}111$ per-skill quality + 3-facet score
calls on the post-dedup survivors (one call per row of the
released \texttt{quality\_judgments} table), $101{,}111$
single-label 16-class classification calls on the same set,
and $66{,}751$ pair-wise dedup adjudication calls on the
borderline embedding-cosine near-duplicate pairs
(\S\ref{sec:pipeline}), per the released dedup report.  Calls were issued at standard worker concurrency
on an internal GPU cluster running
Qwen3.5-397B-A17B-GPTQ-Int4.  The LLM-judge stage dominates
wall time, while content-hash, length-gate, and licence-API stages run
at file-IO / network speed and contribute negligible compute.

\paragraph{Evaluation compute.}
The main grid comprises $24$ configurations ($2$ conditions
$\times$ $4$ cells $\times$ $3$ benchmarks), each run three
times, on the two open Q-27B and
Q-397B backbones, served through Volcano (and, for one cell of
OpenClaw, DashScope).  The Opus~4.7
frontier robustness check ($2$~conditions $\times$ $87$~SkillsBench
tasks $= 174$~task runs) consumed approximately $24$~million
API tokens in total, estimated from the per-task token counters
recorded in the released eval cache
(mean $158$K total tokens per task for the SkillCorpus condition
and $122$K for the no-skill baseline, $87$ tasks each).  The per-task safety-facet regression on
SkillsBench (\S\ref{sec:claim-a}) reuses traces from the main
grid and adds negligible incremental cost.  Aggregate
wall-clock and GPU-hour accounting per ingest pass and per
evaluation cell is provided in the released compute log
alongside the corpus artifact.

% =============================================================================
\subsection{Reproducibility}
\label{app:reproducibility}

The corpus artifact (SQLite database, embeddings, quality cache,
per-stage reject reports) and the ingest pipeline source will be
released upon acceptance, linked from a public Hugging~Face dataset
entry and a GitHub repository hosting the pipeline code under a
permissive licence.

\paragraph{Multi-run merge protocol.}
The released active set is the cumulative union of multiple
ingest passes over the same SQLite database.  Content-hash and
name-hash dedup operate
across all passes, so ingesting a repository whose skills are
content-identical to rows already in the database produces no
duplicate rows.  The released active set is the database snapshot
at release time restricted to rows with $\mathtt{active}=1$ and
$\mathtt{deleted}=0$, the canonical release predicate.  Per-pass
reject counts and the chronological sequence of pipeline
configuration deltas are preserved in the released ingest report.

\paragraph{Reproducing the release predicate.}
A consumer reproducing the released set from the SQLite database
applies the predicate $\mathtt{active}=1 \wedge \mathtt{deleted}=0$
to the \texttt{skills} table, which yields exactly $96{,}401$
rows.  All downstream tables (\texttt{quality\_judgments},
\texttt{source\_priors}, \texttt{vec\_skills}) are keyed on
\texttt{content\_hash} or \texttt{skill\_id} and join on the
released set.

\end{document}